\documentclass[12pt]{iopart}

\usepackage{xcolor}
\usepackage{graphicx}
\usepackage{placeins}
\usepackage{lipsum}
\usepackage{amsfonts}
\usepackage[cases]{cases}
\usepackage{graphicx}
\usepackage{epstopdf}
\usepackage{algorithmic}
\usepackage{cleveref}
\usepackage{booktabs}
\usepackage[tableposition=top]{caption}
\usepackage{subcaption}
\usepackage{array}
\crefname{figure}{Figure}{Figure}
\crefname{table}{Table}{Table}
\ifpdf
  \DeclareGraphicsExtensions{.eps,.pdf,.png,.jpg}
\else
  \DeclareGraphicsExtensions{.eps}
\fi

\crefname{hypothesis}{Hypothesis}{Hypotheses}

\usepackage{amsopn}

\makeatletter
\newcommand*{\addFileDependency}[1]{
  \typeout{(#1)}
  \@addtofilelist{#1}
  \IfFileExists{#1}{}{\typeout{No file #1.}}
}
\makeatother

\usepackage{siunitx}

\DeclareSIUnit\days{days}

\definecolor{darkgreen}{rgb}{0.0, 0.35, 0.0}

\begin{document}
\def\aligned{\vcenter\bgroup\let\\\cr
\halign\bgroup&\hfil${}##{}$&${}##{}$\hfil\cr}
\def\endaligned{\crcr\egroup\egroup}

\title[]{Approximation of the Basset force in the Maxey-Riley-Gatignol equations via universal differential equations}



\author{Finn Sommer$^1$,
  Vamika Rathi$^1$,
  Sebastian Götschel$^1$,
  Daniel Ruprecht$^1$}

\address{$^1$ Chair Computational Mathematics, Institute of Mathematics, Hamburg University of Technology, 21073 Hamburg, Germany}
\ead{finn.sommer@tuhh.de, vamika.rathi@tuhh.de, sebastian.goetschel@tuhh.de, ruprecht@tuhh.de}


\begin{abstract}
The Maxey-Riley-Gatignol equations (MaRGE) model the motion of spherical inertial particles in a fluid.
They contain the Basset force, an integral term which models history effects due to the formation of wakes and boundary layer effects.
This causes the force that acts on a particle to depend on its past trajectory and complicates the numerical solution of MaRGE.
Therefore, the Basset force is often neglected, despite substantial evidence that it has both quantitative and qualitative impact on the movement patterns of modelled particles.
Using the concept of universal differential equations, we propose an approximation of the history term via neural networks which approximates MaRGE by a system of ordinary differential equations that can be solved with standard numerical solvers like Runge-Kutta methods.
In particular, for long-term and high-resolution simulations of trajectories, where the number of required time-steps is large, the UDE approach can reduce time-to-solution.
\end{abstract}

\ack
This project is funded by the Deutsche Forschungsgemeinschaft (DFG, German Research
Foundation) – SFB 1615 – 503850735.

During the preparation and revision of this manuscript, the authors used ChatGPT (OpenAI; including GPT-5.6 Sol during the final revision) for coding support and limited language and LaTeX editing. The authors take full responsibility for the final manuscript and associated code.

%
\vspace{2pc}

%
%
%
%
%
%
%
%
\section{Introduction}
Inertial particles are particles that move immersed in a carrier fluid.
Due to inertial effects, they do not exactly follow the streamlines of the fluid's flow field.
Examples are ubiquitous and include infection transmission~\cite{Viruses}, transport of microplastics in the ocean~\cite{ModelingMicroplatsic} as well as industrial applications \cite{Pharma, IndustrialDust}.

While it is possible to model the movement of inertial particles as a fluid-structure interaction problem where all interactions between fluid and particle, including boundary layers, are fully resolved, such simulations require days or weeks on a high-performance computer.
In situations where the suspension of particles is dilute, the particles are spherical, their center of gravity is aligned with their geometrical center and their size is smaller than the relevant length scales of the flow, their motion can be modeled using the equations of motions proposed by Maxey and Riley~\cite{MaxeyandRiley1983} and Gatignol~\cite{Gatignol1983} in 1983.

These Maxey-Riley-Gatignol equations (or MaRGE for short) are a system of ordinary implicit integro-differential equations.
Their non-dimensional forms reads
\begin{equation}
  \begin{aligned}
    \frac{d\mathbf{y}}{dt}             & =  \mathbf{v}                                                                                                                                                        \\
    \frac{1}{R} \frac{d\mathbf{v}}{dt} & =  \frac{D\mathbf{u}}{Dt} - \frac{1}{S}(\mathbf{v} - \mathbf{u}) - \sqrt{\frac{3}{\pi}\frac{1}{S}} \int^{t}_{t_0}\frac{1}{\sqrt{t - \tau}}(\frac{d\mathbf{v}}{d\tau} - \frac{d\mathbf{u}}{d\tau})  d\tau -(1-R)\mathbf{G},
  \end{aligned}
  \label{eq:MaRGE}
\end{equation}
where $\mathbf{y} \in \mathbb{R}^{3}$ is the position of the particle at time $t$, $\mathbf{v} \in \mathbb{R}^{3}$ is its absolute velocity and $\mathbf{u} \in \mathbb{R}^{3}$ the velocity of the surrounding fluid~\cite{Daitche2013}.
The derivative along the particle trajectory is denoted by $\frac{d\mathbf{u}}{dt}$, while $\frac{D\mathbf{u}}{Dt}$ is the derivative of the corresponding fluid element
\begin{equation}
  \frac{d\mathbf{u}}{dt} = \frac{\partial\mathbf{u}}{dt} + \mathbf{v}\cdot \nabla\mathbf{u} \qquad \text{and} \qquad \frac{D\mathbf{u}}{dt} = \frac{\partial\mathbf{u}}{dt} + \mathbf{u}\cdot \nabla\mathbf{u}.
\end{equation}
Furthermore, $\mathbf{G}$ is the nondimensional gravitational constant, $R$ is a dimensionless parameter $R = \frac{3m_f}{m_f + 2m_p}$ where $m_p$ is the mass of the particle and $m_f$ is the mass of the fluid that the particle displaces.
The Stokes numbere is the non-dimensional parameter $S = \frac{1}{3}\frac{a^2/\nu}{T_{\text{ref}}}$ and includes the characteristic time $T_{\text{ref}}$ of the fluid, the particle radius $a$ and the kinematic viscosity $\nu$.
The larger the Stokes number, the more pronounced is the impact of inertial effects on the movement of the particle~\cite{Bisgaard2020}.
Equation~(\ref{eq:MaRGE}) also requires the initial position $\mathbf{y_0}$ and velocity $\mathbf{v_0}$ of the particle.

Solving~\cref{eq:MaRGE} is not trivial because the integral has a singularity at $t$ and approximating it requires storing every time-step of the simulation.
Because the system is not an ordinary differential equation (ODE), standard solvers like Runge-Kutta methods (RKM) cannot be deployed without modifications.
Analytical solutions exist only for simple fluid fields~\cite{CandelierEtAl2004,PrasathEtAl2019, Rathi2026}.
Because of these difficulties, the Basset force is often ignored, even though its importance has been proven both theoretically~\cite{Daitche_2015, MordantAndPinton2000, Urizarna2025b} and experimentally~\cite{CandelierEtAl2004}.

However, some approaches exist to solve the full MaRGE numerically. 
Daitche et al.~developed a quadrature scheme for the integral term and combine it with explicit Adams-Bashforth multi-step methods of order one to three for the differential part~\cite{Daitche2013}.
By using Tatom's insight that the integral term corresponds to a fractional derivative~\cite{Tatom1988}, Prasath et al.~rewrite the Maxey-Riley-Gatignol equations as a one-dimensional diffusion equation and propose a numerical scheme based on Chebyshev polynomials to solve it~\cite{PrasathEtAl2019}.
Finally, Urizarna-Carasa et al. propose a numerical solver based on finite differences for the reformulated MaRGE~\cite{URIZARNACARASA2025109502}.
While effective, these direct solvers all require a bespoke implementation and cannot rely on existing libraries for solving ODEs.

Other approaches introduce approximations to the history term before solving a simplified version of MaRGE numerically.
Bombardelli et al.~also write the integral as a semi-derivative and approximate it by a series expansion~\cite{BombardelliFractional-Derivative}.
The so-called \textit{window approach} divides the integration time into two parts and splits integral accordingly, exploiting how the impact of the Basset kernel decays over time~\cite{MordantAndPinton2000}.
The first parts integrates the \textit{tail kernel} while the second integral computes the \textit{window kernel} for a recent time interval of fixed length.
The windows kernel is equal to the Basset kernel, while the tail kernel utilizes a computationally more efficient approximation~\cite{MorenoCasasEtAl2016}. 
This approach is motivated by the fact that the Basset kernel decays exponentially after a certain time and becomes, therefore, negligible. 
Van Hinsberg et al.~use exponential functions in the tail kernel \cite{VanHinsbergEtAl2011} whilst González et al.~assume the more distant past can be neglected completely and therefore set the tail kernel to be zero \cite{GonzalezAndrea2007Itpc}.

This paper proposes a data-based approximation of the history term based on universal differential equations (UDEs)~\cite{rackauckas2021}, modelling the Basset force by a universal approximator. 
We compare performance of a standard feedforward neural network (FNN) architecture with that of a long short-term memory (LSTM) with hidden states to better capture the memory effect from the Basset force.
UDEs have been successfully applied to the shallow water equations~\cite{LiuShallowWater}, systems biology~\cite{Philipps2025} and neuroscience~\cite{El-GazzarNeuroscience} but have not yet been investigated for MaRGE.
By replacing the integral with a neural network, we reduce the complexity of solving MaRGE to that of a standard ordinary differential equation for which effective and easily usable libraries exist.
Throughout the paper, we use the midpoint rule, implemented in the \textit{DifferentialEquations} package in Julia~\cite{rackauckas2017differentialequations}, but any other method provided by the library could be used easily.
All results shown in the paper can be reproduced using the following published codes for data-generation~\cite{flsommer_numerical}, training~\cite{flsommer_udes} and evaluation~\cite{flsommer_analysis}, as well as the published data~\cite{flsommer_data}.

%
%
%
%
%
%
%
%
\section{Data generation: flow fields and full MaRGE trajectories}
 First, \S\ref{sec:daitche} describes the numerical solver by Daitche that we use to solve the full MaRGE to generate training data.
Then, \S\ref{subsec:flowfields} describes the two flow fields in which we compute inertial particle trajectories to test our approach.
The first field is a 3D analytical vortex-field with a time- and $z$-dependent angular frequency.
The second field is a 3D field that has been interpolated from measurement data obtained from a lab-sized stirred tank reactor at the University of Applied Sciences Hamburg (HAW)~\cite{WEILAND2023100448}.

%
%
%
%

\subsection{Daitche's method and trajectory generation} \label{sec:daitche}
To solve MaRGE,~\cref{eq:MaRGE}, we use the method developed by Daitche~\cite{Daitche2013} and extended to 3D by Rathi~\cite{Rathi2026} where the history term is approximated with specially designed quadrature schemes combined with an Adams-Bashforth multistep method of orders one, two or three for the differential part.
Throughout this paper, we use the third order variant in all cases.

When recast for the particle's relative velocity $\mathbf{w} = \mathbf{v} - \mathbf{u}$, MaRGE becomes
\begin{equation}\label{eq:MaRGE2}
	\frac{\mathrm{d}\mathbf{w}}{\mathrm{d}t} =
	(R-1)\,\frac{\mathrm{d}\mathbf{u}}{\mathrm{d}t}
	- \frac{R}{S}\mathbf{w}
	-R \mathbf{w}\cdot \nabla \mathbf{u}
	- R\sqrt{\frac{3}{S\pi}}
	\frac{\mathrm{d}}{\mathrm{d}t}
	\int_{t_0}^{t} \frac{\mathbf{w}(\tau)}{\sqrt{t-\tau}}
	\,\mathrm{d}\tau
	- (1-R)\mathbf{G}.
\end{equation} 
Setting
\begin{equation}
	\mathbf{\tilde{G}} := 
	(R-1)\,\frac{\mathrm{d}\mathbf{u}}{\mathrm{d}t}
	- \frac{R}{S}\mathbf{w}
	-R \mathbf{w}\cdot \nabla \mathbf{u}
	- (1-R)\mathbf{G},
\end{equation}
\begin{equation}
	\mathbf{H} := 
	- R\sqrt{\frac{3}{S\pi}}
	\int_{t_0}^{t} \frac{\mathbf{w}(\tau)}{\sqrt{t-\tau}}
	\,\mathrm{d}\tau,
\end{equation}
\cref{eq:MaRGE2} becomes
\begin{equation}
	\frac{\mathrm{d}\mathbf{w}}{\mathrm{d}t} = 
	\mathbf{\tilde{G}} + \frac{\mathrm{d}}{\mathrm{d}t} \mathbf{H}.
\end{equation}
We divide the time domain into time steps $(t_n, t_n + h)$ of equal length $h$. 
By integrating from $t_n$ to $t_n+h$ we get
\begin{equation}\label{int_MaRGE}
	\mathbf{w}(t_n + h) = \mathbf{w}(t_n) + \int_{t_n}^{t_n + h} \mathbf{\tilde{G}}(\tau) \mathrm{d}\tau + 
	\mathbf{H}(t_n + h) - \mathbf{H}(t_n).
\end{equation}
The value of $\mathbf{H}$ at $t_n + h$ is calculated by Daitche~\cite{Daitche2013} using a quadrature scheme
\begin{equation}\label{eq:H_calc}
	\mathbf{H}(t_n + h) = -\xi \sum_{j=0}^{n+1} \mu_{j}^{n+1} \mathbf{w}(t_{n+1-j}) + \mathcal{O}(h^m),
\end{equation}
where the values of the coefficients $\mu_{j}^{n+1}$ depend on $\xi = \sqrt{h} \, R \sqrt{3/(S \pi)}$.
The approximation of $\int_{t_n}^{t_n + h} \mathbf{\tilde{G}}(\tau) \mathrm{d}\tau$ is obtained by an Adams-Bashforth multi-step method.
Letting $\mathbf{w}_n = \mathbf{w}(t_n)$, the complete integration scheme for~\cref{int_MaRGE} is
\begin{equation}
	\mathbf{w}_{n+1} = \mathbf{w}_n + \int_{t_n}^{t_n + h} \mathbf{\tilde{G}}(\tau) \mathrm{d}\tau - \xi \sum_{j=0}^{n+1} \mu_{j}^{n+1} \mathbf{w}_{n+1-j} + \xi \sum_{j=0}^{n} \mu_{j}^{n} \mathbf{w}_{n-j},
\end{equation}
which is then solved for $\mathbf{w}_{n+1}$ so that
\begin{equation}
	(1+\xi \mu_{0}^{n_1}) \mathbf{w}_{n+1} = \mathbf{w}_n + \int_{t}^{t+h} \mathbf{\tilde{G}}(\tau) \mathrm{d}\tau - \xi \sum_{j=0}^{n} (\mu_{j+1}^{n+1} - \mu_{j}^{n}) \mathbf{w}_{n-j}.
\end{equation}
The increment in the history term can be calculated from
\begin{equation}
	\mathbf{H}(t_n + h) - \mathbf{H}(t_n) = -\xi \mu_{0}^{n+1} \mathbf{w}_{n+1} - \xi \sum_{j=0}^{n} (\mu_{j+1}^{n+1} - \mu_{j}^{n}) \mathbf{w}_{n-j},
\end{equation}
so that $\mathbf{H}$ at every grid point can be determined recursively by setting $\mathbf{H}(t_0) = \mathbf{0}$.
%
%
%
\subsection{Flow Fields} \label{subsec:flowfields}
\begin{figure}[th]
  \centering
  \begin{subfigure}[]{0.49\textwidth}
    \centering
    \includegraphics[width = \textwidth]{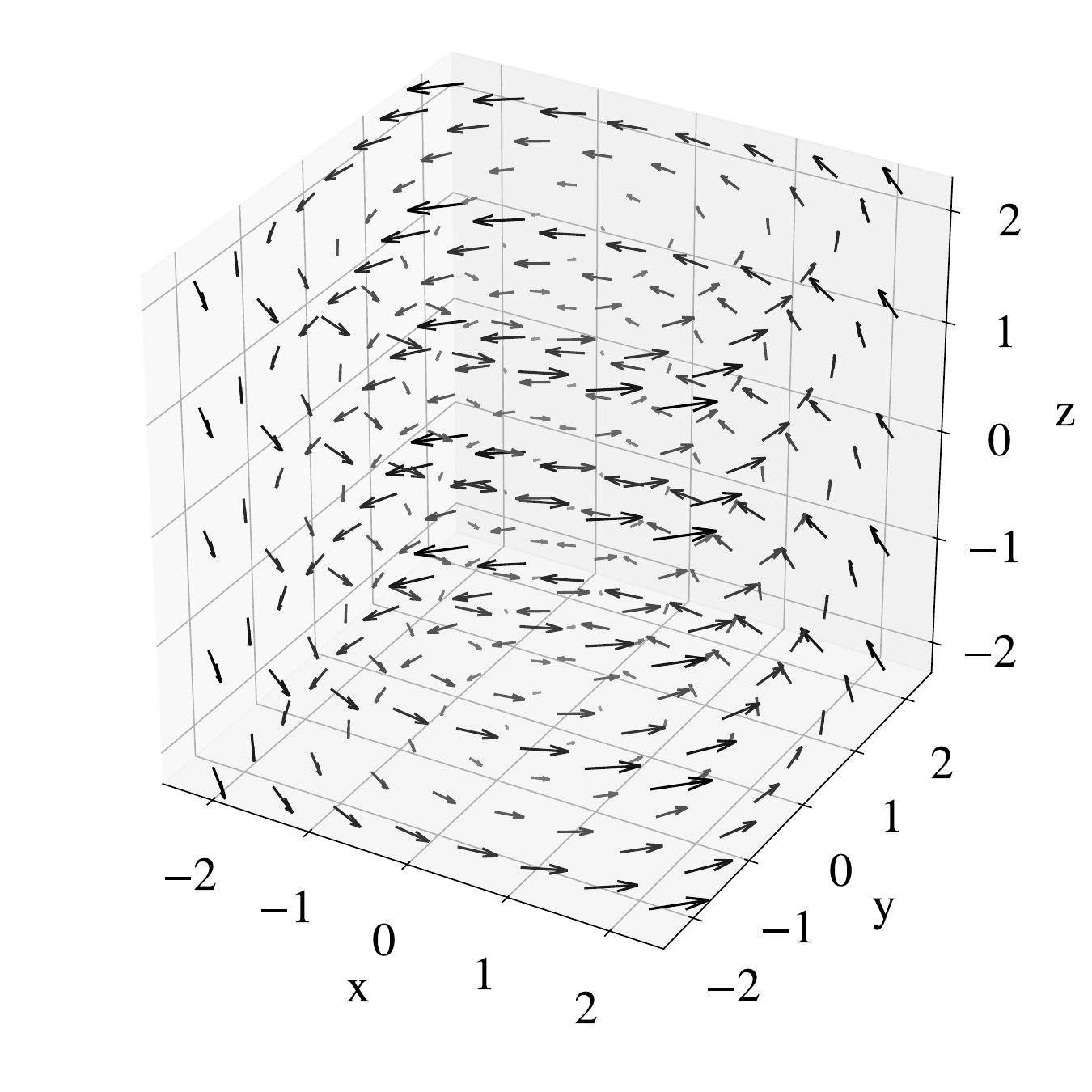}
    \caption{Analytical flow field.}
  \end{subfigure}
  \begin{subfigure}[]{0.49\textwidth}
    \centering
    \includegraphics[width = \textwidth]{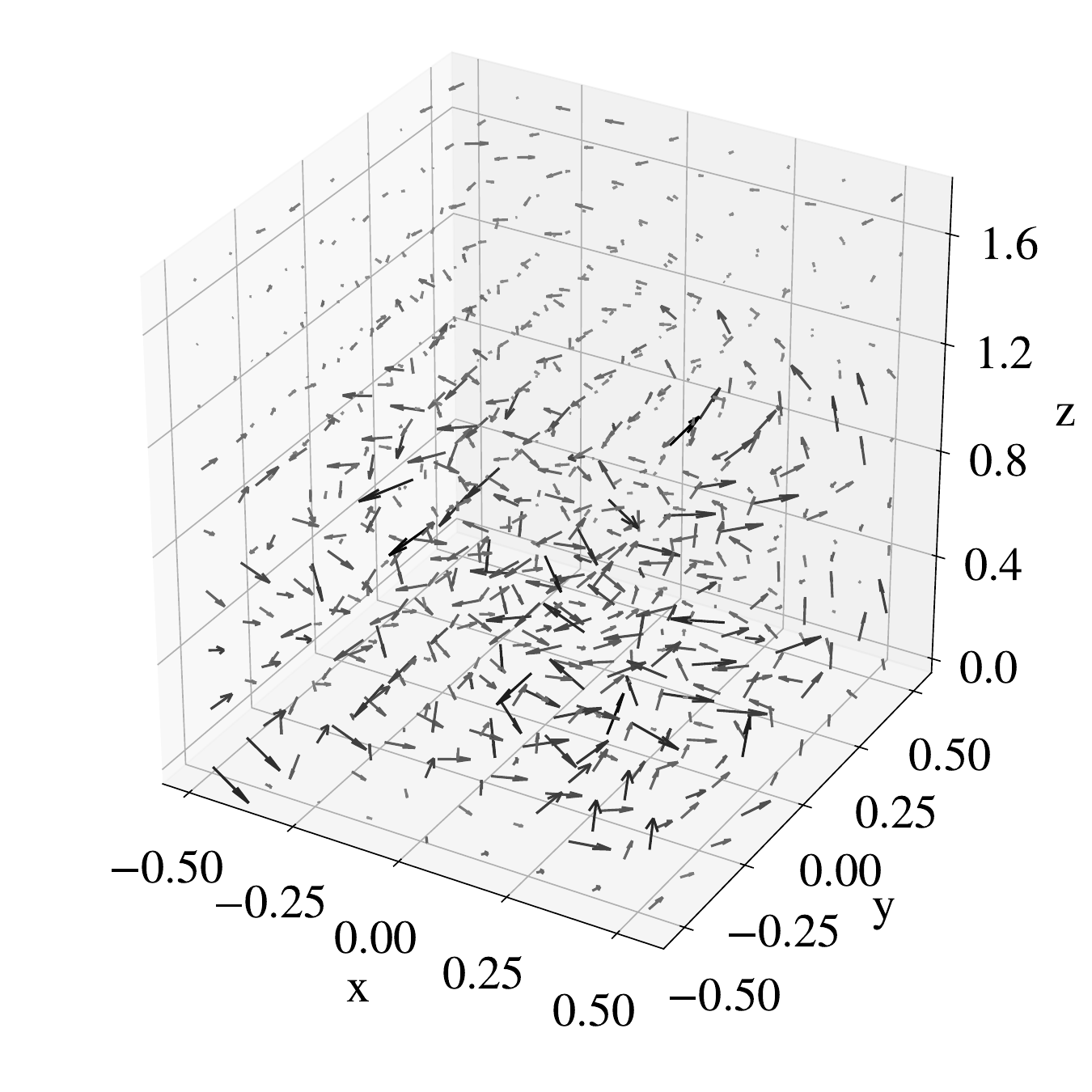}
    \caption{Experimental flow field.}
  \end{subfigure}
  \caption{Quiver plots of the two fluid fields used in the numerical experiments.}
  \label{fig:FluidFields}
\end{figure}

Our first flow field
\begin{equation}
  \vec{u}(x, y, z, t) =
  \left[
    \begin{array}{c}
      -y  \omega(z, t) \\
      x  \omega(z, t)  \\
      0.0
    \end{array}
    \right].
  \label{eq:vortex}
\end{equation}
is based on the 2D vortex fluid field \cite{CandelierEtAl2004} but extended to 3D by letting the angular frequency $\omega$ depend on $z$ and $t$ and adding gravity to MaRGE.
Note that there is no fluid velocity in the vertical direction but in all examples we consider particles that are denser than the fluid so that buoyancy causes downward vertical motion.
We use a $z$- and $t$-dependent angular frequency
\begin{equation}
  \omega(z, t) = \omega_0 +  \alpha \sin^2(z) \cos^2(t),
\end{equation}
with $\omega_0 = 1.0$ and $\alpha = 0.2$.
A quiver plot of the field at $t = 0.0$ is shown in~\cref{fig:FluidFields} (left). 
The magnitude of the velocity vector is indicated by the arrow length.

The second flow field is interpolated from data obtained from a lab-scale stirred tank reactor of 2.8 litres with a Korbbogen head bottom, three baffles, and two Rushton turbines~\cite{WEILAND2023100448}.
Over a period of two seconds, the experiment uses high-speed cameras to take a large number of pictures of passive tracer particles inside the tank.
Trajectories are computed from these pictures using the \textit{shake-the-box} method~\cite{Schanz2016} to provide a Lagrangian description of the fluid flow.
This is mapped to a Eulerian grid via a scattered interpolant algorithm implemented in Matlab.

This Eulerian data was used in the experiments for this paper.
It contains values for the three components of $\mathbf{u}$ at $62 \times 62 \times 107$ grid points over a time of two seconds with a frequency of $0.002$ per second for a total of 1001 snapshots.
The interpolated grid-data is of a size of $(x, y, z) \in [-6.386 \cdot 10^{-2}; 6.925 \cdot 10^{-2}] \times [-6.638 \cdot 10^{-2}; 6.739 \cdot 10^{-2}] \times [2.4046 \cdot 10^{-3}; 2.289 \cdot 10^{-1}]$ meters.
For memory efficiency and noise suppression, we sparsen the data by averaging the three first and the three last time-steps, as well as five consecutive time-steps in between, leaving a total of 201 snapshots. 
All lengths, times and velocities are non-dimensionalized using the reference values in~\cref{tab:characteristic_values}.

\begin{table}
  \centering
  \begin{tabular}{l c} \toprule
			Param.      & Value  \\
			\midrule
			$u_{\text{ref}}$         & $4.8 \cdot10^{-1}\frac{m}{s}$ \\
			$T_{\text{ref}}$          & $2.7 \cdot10^{-1}s$  \\
			$L_{\text{ref}}$   & $1.296 \cdot10^{-1}m$  \\ \bottomrule
		\end{tabular}
  \caption{The values of the characteristic -velocity, -time, and -length of the experimental fluid field used to non-dimensionalize it for the MaRGE. We set the length scale such that $L_{\text{ref}} = u_{\text{ref}} / T_{\text{ref}}$ to maintain consistency.}
  \label{tab:characteristic_values}
\end{table}

To solve MaRGE numerically, we need to evaluate $\mathbf{u}$ and its material derivative at arbitrary points $(x, y, z, t)$. 
In time, we use linear interpolation so that 
\begin{equation*}
    \hat{\mathbf{u}}(x,y,z,t) = \hat{\mathbf{u}}(x,y,z,t_i) + \frac{\hat{\mathbf{u}}(x,y,z,t_{i+1}) - \hat{\mathbf{u}}(x,y,z,t_i)}{t_{i+1} - t_{i}}(t - t_i)
\end{equation*}
for $t_i \leq t \leq t_{i+1}$ with $t_i$, $t_{i+1}$ being the times of the closest available snapshots.
Correspondingly, the time-derivative is approximated by
\begin{equation}
	\frac{\partial \mathbf{u}(x, y, z, t)}{\partial t} \approx \frac{\hat{\mathbf{u}}(x,y,z,t_{i+1}) - \hat{\mathbf{u}}(x,y,z,t_i)}{t_{i+1} - t_{i}}.
\end{equation}
In space we use cubic B-spline interpolation using the interpolate function from the \textit{interpolation.jl} package in Julia \cite{Interpolationsjl}, whilst derivatives are computed using finite differences.

%
%
%
%
%
%
%
%
\section{Universal Differential Equations}
 Universal Differential Equations (UDEs) connect physical modelling via differential equations with data-driven approaches like machine learning~\cite{rackauckas2021}. 
UDEs replace individual terms of a differential equation with a universal approximator, typically a neural network.
These are then trained to learn unknown dynamics or to use data to compensate for model inaccuracies or limitations. 
We replace the Basset term in~\cref{eq:MaRGE2} by a neural network $\mathcal{NN}$ and approximate MaRGE by
\begin{equation}
  \begin{aligned}
    \frac{d\mathbf{y}}{dt} & =  \mathbf{v} = \mathbf{w} + \mathbf{u}                                                                                                        \\
    \frac{d\mathbf{w}}{dt} & =  (R -1)\frac{d\mathbf{u}}{dt} - \frac{R}{S}\mathbf{w} - R \mathbf{w} \cdot \nabla \mathbf{u}                                                 \\
                           & - R\sqrt{\frac{3}{S\pi}}\mathcal{NN}(t, \mathbf{q}, \frac{d\mathbf{w}}{dt}, \mathbf{u}, \frac{D\mathbf{u}}{Dt}, \mathbf{u_0}) - (1-R)\mathbf{G}.
  \end{aligned}
  \label{eq:UDEMaRGE}
\end{equation}
Below, we will experiment with either a feedforward neural network (FNN) or an LSTM with hidden states.
The network input consists of the time $t$, the approximate state of the particle at this time $\mathbf{q} \in \mathbb{R}^6$, the derivative of the solution $\mathbf{\frac{dw}{dt}} \in \mathbb{R}^3$, 
the fluid velocity $\mathbf{u}$, the material derivative of the fluid field $\mathbf{\frac{Du}{Dt}}$, and the fluid velocity at the initial position of the particle $\mathbf{u_0}$.
Note that state $\mathbf{q} = (\mathbf{y}, \mathbf{w}) \in \mathbb{R}^{6}$ of the particle consists of its position and velocity so that the network has a total of 19 input parameters.  
To match the dimensionality of the equation, the network always has to have three output parameters.

For some initial particle state $\mathbf{q}_0 = (\mathbf{y_0}, \mathbf{w_0}) \in \mathbb{R}^{6}$, \cref{eq:UDEMaRGE} can be solved using a standard ODE integrator.
Here, we use the second order, A-stable midpoint rule.
In all cases, the initial relative velocity of the particle is set to zero.

%
%
%
%

\subsection{Network architectures}
The first architecture we explore is a feedforward network (FNN)~\cite{SCHMIDHUBER201585} with four hidden layers, each containing 64 neurons.
Including the biases for each layer, this results in a total of 13955 trainable parameters.
The chosen activation function is the $\tanh$-function except for the output layer, where no activation function is applied to not restrict the solution space.
The second model is a combined model, consisting of an LSTM cell~\cite{HochreiterLSTM} and a dense output layer. 
LSTMs are a recurrent type of network architecture that provides the ability to also consider past inputs, matching to some degree the workings of the integral term.
It has two states: the \textit{cell state}, which acts as the long-term memory, and a \textit{hidden state}, which acts simultaneously as a short-term memory and as the output of the LSTM-cell.
\begin{figure}[t]
  \centering
  \includegraphics[width=\textwidth]{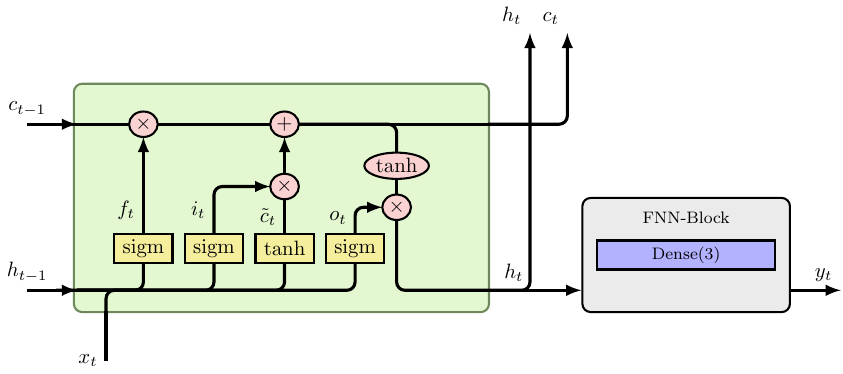}
  \caption[LSTM-Architecture]{The used LSTM-Architecture consists of a recursive LSTM-cell that utilizes a \textit{cell state} $c_t$ and a \textit{hidden state}
  $h_t$ to store information about previous timesteps. A FNN-Block with a single layer of three neurons is connected to the output of the LSTM to increase the hidden state to arbitrary size.}
  \label{fig:LSTM-Architecture}
\end{figure}
The number of trainable parameters in the LSTM is only determined by the input and output dimensions. 
Since we need the output to be three-dimensional, it is therefore not possible to add more parameters to the LSTM cell itself, in contrast to the FNN which can be made deeper and wider as the user desires.

To give the LSTM more flexibility and match its size to the FNN, we add a dense layer after the output of the LSTM model. 
The model is written using the Lux framework in Julia~\cite{pal2023lux,pal2023efficient}. The template for the LSTM comes from the Lux documentation, where model with that structure is applied to classify clockwise and anticlockwise spirals~\cite{pal2023lux}. 
This dense layer consists of three neurons and allows to increase the hidden state to an arbitrary size.
The hidden state consists of 48 neurons, which leads to a total of 13395 trainable parameters, very slightly fewer than the FNN.
Logistic/sigmoid and tanh function are used as activation functions of the LSTM, which is the default choice~\cite{LSTM-activations}. 
The output network uses no activation function.
A sketch of the LSTM architecture is shown in~\cref{fig:LSTM-Architecture} and a summary of the most important parameters of both architectures is available in~\cref{tab:Overview networks}.
\begin{table}[th]
	\setlength{\tabcolsep}{10pt}
  \centering
  \begin{tabular}{l l l} \toprule
    Model Name           & FNN          & LSTM                       \\
    \midrule
    activation functions & $\tanh$      & logisitc function, $\tanh$ \\
    Recurrent            & No           & Yes                        \\
    Trainable params     & 13955        & 13395                      \\
    Size parameter file  & $122.8 $ KiB & $117.0$ KiB                \\ \bottomrule
  \end{tabular}
  \caption{Overview of the network parameters for both network architectures. The size corresponds to the \texttt{jld2} file that is generated when saving all parameters.}
  \label{tab:Overview networks}
\end{table}
\subsection{Training Setup}
In order to provide training data for the networks, we numerically solve~\cref{eq:MaRGE} using the solver described in~\cref{sec:daitche}.
Initial positions are sampled randomly within a fixed area while initial relative velocities are always set to zero.
Important parameters for both setups can be found in~\cref{tab:overview parameters}.

For both the vortex and experimental field, we generate three types of trajectory sets.
The first is the training set which contains all data used to optimize the network's weights. 
The second is the validation set used to evaluate the model during training and identify overfitting.
Validation data is not used to optimize the network's weight and bias parameters but to monitor generalization capabilities and to adjust hyper-parameters such as the learning rate. 
For each initial position in the training and validation set, the ground-truth trajectory is computed over a time-interval $T_{\text{train}}$, see~\cref{tab:overview parameters} for the concrete values.

The third dataset is the test set. 
Trajectories within this set are not touched during training and are only used post-training to evaluate the network's ability to generalize.
FNN and LSTM are trained, evaluated and tested on the same data sets.
In contrast to training and validation, trajectories for initial positions in the test set are computed over the longer time interval $T_{\text{test}}$, see~\cref{tab:overview parameters}.
This allows to test how the UDE generalizes beyond the training interval.

For the vortex field,~\cref{eq:vortex}, we perform six experiments with increasingly larger training sets containing  $1, 5, 10, 25, 50$, and $100$ trajectories.
The sets are nested so that the five trajectories in the second experiment are part of the 10 trajectories in the third and so on.
The aim is to gain  insight into how the size of the training data set correlates with the precision of the approximation.
Initial positions are sampled randomly in the unit cube $[-1, 1]^3$.
The validation set consists of 20 trajectories, also sampled randomly from the unit cube, which remain the same for all six experiments.

\begin{table}
			\setlength{\tabcolsep}{10pt}
	\begin{minipage}{0.49\linewidth}
		\centering
		\begin{tabular}{c c} \toprule
			Param.      & Value  \\
			\midrule
			$R$         & $0.968$ \\
			$S$         & $1.0$  \\
			$T_{\text{train}}$ & [0.0, 10.0]   \\
			$T_{\text{test}}$ & [0.0, 60.0]   \\
			$h$         & 0.1    \\ \bottomrule
		\end{tabular}
		
		\subcaption{Vortex field}
		\label{tab:Parameter MaRGE Vortex field}
	\end{minipage}
	\begin{minipage}{0.49\linewidth}
		\centering
	\begin{tabular}{c c} \toprule
		Param.   & Value               \\
		\midrule
		$R$         & $0.968$ \\
		$S$         & $1.23$  \\
		$T_{\text{train}}$ & [0.0, 1.852]   \\
		$T_{\text{test}}$ & [0.0, 7.407]  \\
		$h$         & 0.01    \\ \bottomrule
	\end{tabular}
		\subcaption{Experimental field}
		\label{tab:Parameter MaRGE Experimental field}
	\end{minipage}
	\caption{Dimensionless and numerical parameters used in the experiments.}
	\label{tab:overview parameters}
\end{table}

For the experimental field we conduct a single experiment with 50 trajectories in the training set. 
Dimensional initial positions are sampled within a domain
$[-2 \cdot 10^{-2}, 2 \cdot 10^{-2}] \times [-2 \cdot 10^{-2}, 2 \cdot 10^{-2}] \times [1\cdot 10^{-1}, 2\cdot 10^{-1}]$ such that $x-$ and $y-$ coordinates are near the center of the reactor and the $z-$ coordinate in its upper part. 
Sampled values are then non-dimensionalized using the reference length in~\cref{tab:characteristic_values}.
Since we do not consider any interactions with the reactor's walls, this prevent too many simulated particles moving out of the region where data is available.
Trajectories that still leave the domain are discarded and not used for the UDE training.
The validation set consists of 20 trajectories whose initial positions are sampled within the same area as the training set. 
Both validation and training trajectories are integrated over the time-interval $T_{\text{train}}$, see~\cref{tab:overview parameters}.

We use the average squared difference between the ground-truth trajectory generated by solving the full MaRGE~\cref{eq:MaRGE2} and the trajectory from integrating the UDE system~\cref{eq:UDEMaRGE} as loss function.
Let $(t_1, t_2, ..., t_n)$ with $t_i \in \mathbb{R}$ denote the $n$ time-steps of a pair of trajectories, which are chosen to be identical for both ground-truth and UDE.
Let $Q, \hat{Q} \in \mathbb{R}^{n \times 6}$ be the particle states at all time-steps, predicted by solving MaRGE or the UDE system, with each state $q_i, \hat{q}_i$ consisting of the particle's position $\mathbf{y}_i \in \mathbb{R}^3$ and relative velocity $\mathbf{w}_i \in \mathbb{R}^3$ at time $t_i$.
Given a pair consisting of a ground-truth trajectory $Q$ and an approximation $\hat{Q}$, the loss function $\mathcal{L}$ is
\begin{equation}
  \mathcal{L}(Q, \hat{Q}) = \frac{1}{n}\sum_{i = 1}^{n} (||q_i - \hat{q}_i||_2^2).
\end{equation}
Training proceeds in two phases. 
The Adam optimizer~\cite{AdamPaper} is applied the first 300 epochs.
After that, the L-BFGS optimizer~\cite{Liu1989-LBFGS} is used to further reduce training losses for 200 iterations.
For the vortex field, the learning rate is 0.01 for both FNN and LSTM.
For the experimental field, the learning rate for the LSTM remains 0.01 but, to improve stability of the training process, the FNN learning rate is reduced to 0.001.

%
%

%
\subsection{Training Results}
In this section, we discuss the results of the training process for the two architectures and flow fields.

\begin{figure}[t]
  \centering
  \includegraphics[width=\textwidth]{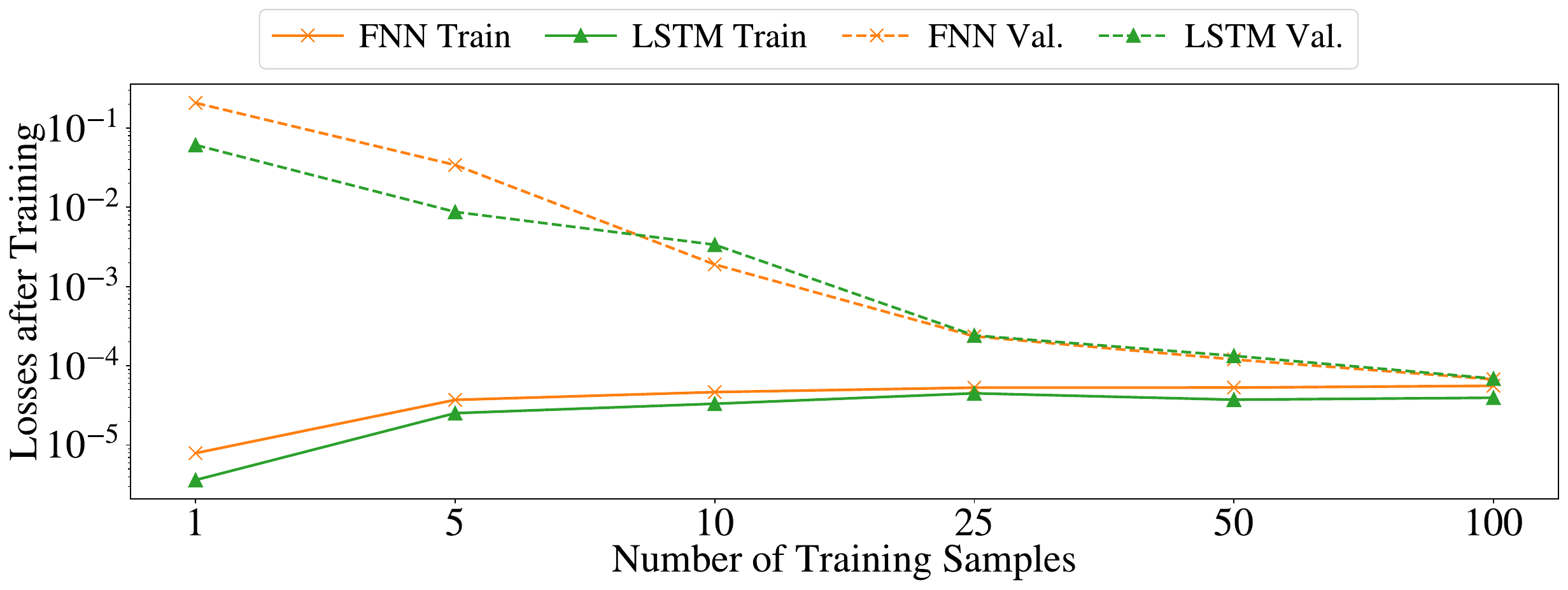}
  \caption{Final training and validation losses at the end of training depending on the number of trajectories in the training set. A relatively modest number of 25 reference trajectory pairs is sufficient to push training and validation losses below $10^{-3}$} for the vortex field. The LSTM produces slightly smaller losses for small training data sets, but the difference is minor.
  \label{fig:Loss Convergence first Experiment}
\end{figure}

\paragraph{Vortex field.}
The final loss at the end of training versus the number of trajectories in the training set is shown in~\cref{fig:Loss Convergence first Experiment}.
For the largest training data set with 100 trajectories, both networks show similar final training losses. 
The final loss of the LSTM of $3.96\cdot10^{-5}$ is slightly below that of the FNN with $5.603\cdot10^{-5}$, but the difference is marginal.
For smaller training data sets, in particular with five trajectories, the LSTM produces slightly lower losses than the FNN.
While this indicates some advantage from the hidden states when it comes to learning the history effect from the Basset force, the effect remains minor.

\begin{figure}[t]
  \centering
  \includegraphics[width=\textwidth]{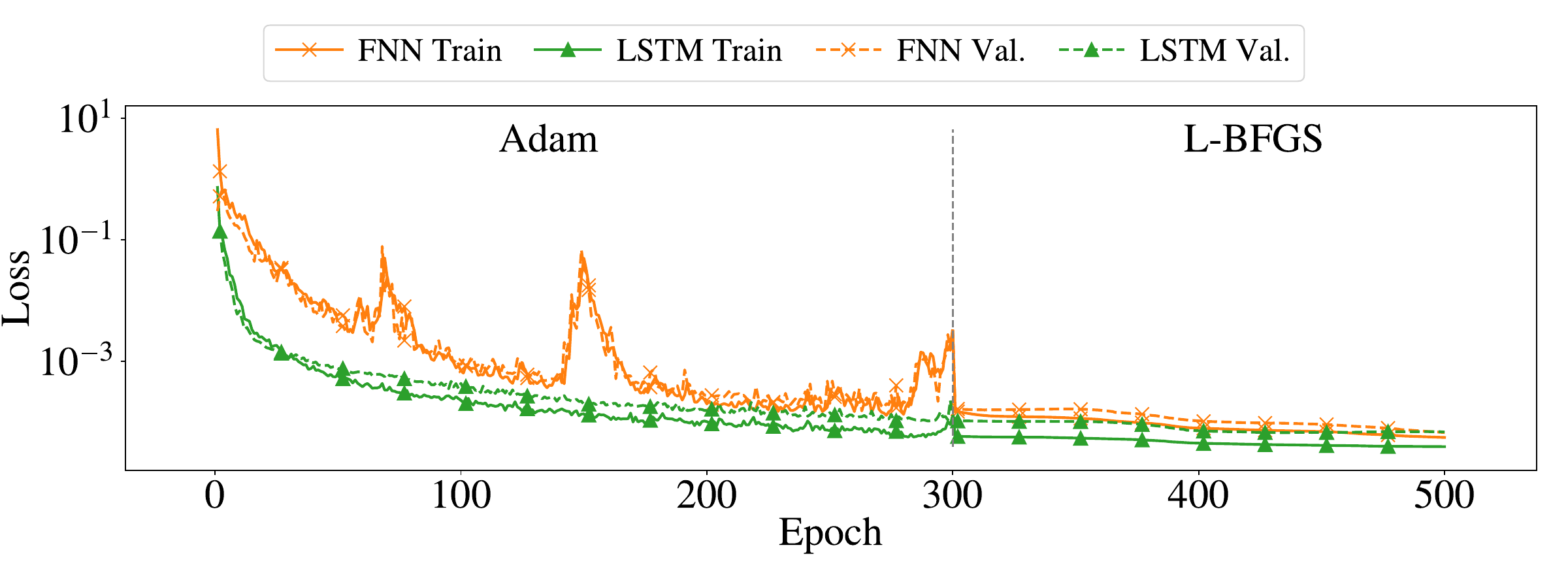}
  \caption{Training and validation loss during training for a training set with 100 trajectories for the vortex field. While the LSTM has some advantage in early stages of training, both architectures produce very similar losses when trained long enough.}
  \label{fig:Loss_curve}
\end{figure}

In~\cref{fig:Loss_curve} we show training and validation losses over the course of the training for the case of 100 trajectories in the training set.
Both training and validation losses are lower for the LSTM than for the FNN, and a lot less volatile.
At the end of the first training phase at epoch 300, the LSTM provides a smaller validation loss of $1.292\cdot10^{-4}$ compared to $3.64\cdot10^{-3}$ for the FNN. 
However, the higher loss for the FNN is mostly an artefact of its volatility.
At epoch 263, the validation loss of the FNN is $1.53\cdot10^{-4}$ and therefore comparable to the LSTM.
The second training phase using L-BFGS produces only a small further reduction of losses.
In summary, while the LSTM has some advantages if training is stopped very early, both networks deliver very comparable accuracy when trained sufficiently long.

\begin{figure}[t]
  \centering
  \includegraphics[width=\textwidth]{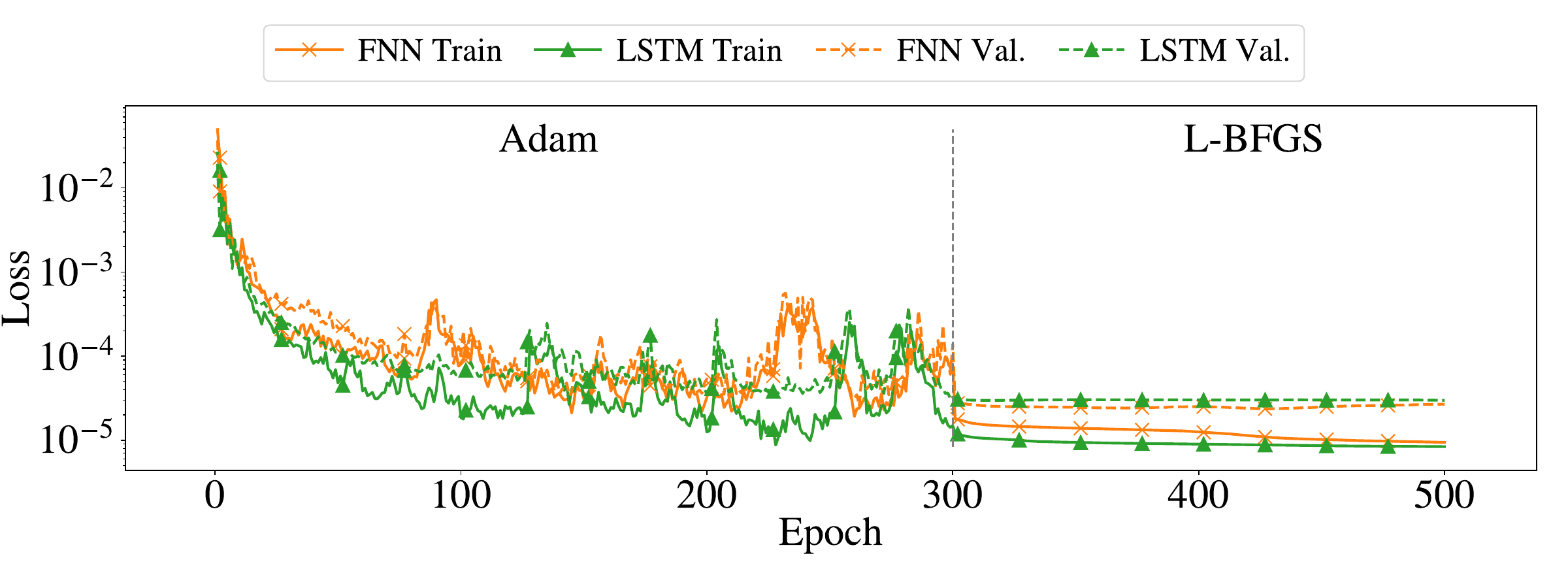}
  \caption{Training and validation loss for the experimental field. The LSTM retains an advantage in early training but also produces somewhat lower final losses than the FNN toward the end of training, in contrast to the vortex field.}
  \label{fig:Loss_curve_exp02}
\end{figure}

\paragraph{Experimental field.}
Training and validation losses for the experimental field are shown in~\cref{fig:Loss_curve_exp02}.
Overall progression of training is very similar to the vortex field but the LSTM loss is somewhat more volatile.

At the end of the first training phase, the training losses are $7.845\cdot10^{-5}$ for the FNN and $1.473\cdot10^{-5}$ for the LSTM, whilst validation losses are $1.392\cdot10^{-4}$ and $3.325\cdot10^{-5}$. 
The second training phase with L-BFGS reduces training losses further to $9.408\cdot10^{-6}$ for the FNN and $8.371\cdot10^{-6}$ for the LSTM.
Validation losses behave similarly and reduce to $2.672\cdot10^{-5}$ for the FNN and $2.971\cdot10^{-5}$ for the LSTM at the end of the second training phase, illustrating that both models perform equally well.
Overall, both networks train successfully with reductions of training and validation losses by five orders of magnitude.

\subsection{Computational cost}
In this section, we compare the full Daitche solver and the UDE approach with respect to memory footprint and runtime for time-step sizes of $h=0.1$ and $h=0.01$ over a time interval $[0,60]$. 
We use Daitche's third-order solver based on Adams-Bashforth 3 as the integrator for the full MaRGE and both AB-3 and the midpoint rule as integrators for the UDEs.
Note that the quadrature strategy used to approximate the integral term in Daitche's method cannot easily be adopted for implicit methods.
For a number of time-steps $N = \mathcal{O}(1/h)$, the UDE approach, using a standard solver for an ODE, scales like $\mathcal{O}(N)$ with respect to computational cost while the memory footprint is $\mathcal{O}(1)$, because older time-steps can be discarded or stored to disk.
By contrast, Daitche's method requires storage of all time-steps as well as of $\mathcal{O}(N^2)$ many quadrature weights.
The convolution structure of the memory term also requires $\mathcal{O}(N^2)$ operations per time-step.
Therefore, we expect the UDE to deliver shorter time-to-solution and require less memory as the resolution $h$ becomes smaller.

To confirm this, we calculate the same trajectory 100 times and measure solution times using the Julia macro \texttt{@timed}, excluding setup times.
\cref{tab:computational_resources} shows that for $h=0.1$, Daitche's method is still substantially faster than the UDEs.
However, decreasing $h$ by a factor of ten to $h=0.01$ increases runtimes of Daitche's solver by about two orders of magnitude while the cost for the UDEs increases only by one order. 
Therefore, the UDE method now delivers shorter time-to-solution, and the gains will get more pronounced if $h$ is reduced further or a longer trajectory is computed.

In terms of memory demand, we  use \texttt{Sys.maxrss()} to measure the Peak Resident Set Size (Peak RSS) of the Julia process calculating one trajectory and average over five independent runs.
However, note that this measures all memory demands, including those from the compilation process and the loaded packages needed by the \texttt{SciML} framework, which are not required by the Daitche solver.
Therefore, the UDE's memory footprint is about double that of Daitche. 
Some Julia packages exist that allow for the stateless saving of network weights and more memory-efficient inference, but studying this is beyond the scope of this paper.
To give a more precise estimate of the memory demands, we use \texttt{Base.summarysize()} to measure the memory required to store the integration weights for Daitche's solver or the network weights for the UDE using the Julia built-in function \texttt{Base.summarysize()}.
Going from $h=0.1$ to $h=0.01$ increases memory demand for Daitche's solver by two orders of magnitude, in line with the quadratic scaling explained above, while memory demand for the UDE does not change as the resolution is refined.

\begin{table}
    \centering
    \setlength{\tabcolsep}{6pt}
    \small

    \begin{tabular}{l c c c c c}
        \toprule
        & Daitche
        & \multicolumn{2}{c}{UDE--FNN}
        & \multicolumn{2}{c}{UDE--LSTM} \\
        \cmidrule(lr){2-2}
        \cmidrule(lr){3-4}
        \cmidrule(lr){5-6}
        & AB3
        & AB3 & Midpoint
        & AB3 & Midpoint \\
        \midrule

        \multicolumn{6}{c}{\textit{Trajectory with $h=0.1$}} \\
        \midrule
        time [s]
            & $5.4 \cdot 10^{-4}$ & $2.15 \cdot 10^{-3}$ & $3.94 \cdot 10^{-3}$ & $2.11 \cdot 10^{-3}$ & $3.95 \cdot 10^{-3}$ \\
        Peak RSS [MiB]
            & $4.36\cdot 10^{2}$ & $9.81\cdot 10^{2}$ & $9.76 \cdot 10^{2}$ & $1.05 \cdot 10^{3}$ & $1.04 \cdot 10^{3}$ \\
        Method coefficient [MiB]
            & $3.1 \cdot 10^{0}$ & $1.07 \cdot 10^{-1}$ & $1.07 \cdot 10^{-1}$ & $1.02 \cdot 10^{-1}$ & $1.02 \cdot 10^{-1}$ \\

        \midrule
        \multicolumn{6}{c}{\textit{Trajectory with $h=0.01$}} \\
        \midrule
        time [s]
            & $1.04 \cdot 10^{-1}$ & $1.95 \cdot 10^{-2}$ & $3.76 \cdot 10^{-2}$ & $2.2 \cdot 10^{-2}$ & $4.71 \cdot 10^{-2}$ \\
        Peak RSS [MiB]
            & $7.89 \cdot 10^{2}$ & $1.02\cdot 10^{3}$ & $1.03 \cdot 10^{3}$ & $1.14 \cdot 10^{3}$ & $1.13 \cdot 10^{3}$ \\
        Method coefficients [MiB]
            & $2.75\cdot 10^{2}$ & $1.07 \cdot 10^{-1}$ & $1.07 \cdot 10^{-1}$ & $1.02 \cdot 10^{-1}$ & $1.02 \cdot 10^{-1}$ \\

        \bottomrule
    \end{tabular}

    \caption{Computational resources required by the Daitche and UDE solvers 
    for two time-step sizes. Mean time is the averaged wall-clock time required to compute one trajectory. Peak RSS is the maximum amount of memory required for this computation, while ``Method coefficient'' measures only the memory required to store the quadrature weights in Daitche's case or the network weights for the UDE.}
    \label{tab:computational_resources}
\end{table}

%
%
%
%
%
%
%
%
\section{Testing and generalization}
After training, the models are used to generate the test trajectories over the longer time-interval $T_{\text{test}}$.
A total of 125 trajectory pairs are generated from equally spaced initial on a $5 \times 5 \times 5$ grid convering the same area as the training trajectories.
For the experimental field, every trajectory that leaves the domain is discarded, which leaves a total of 99 trajectory pairs. 
All results reported in this section are for trajectories in the test set which were not considered in any way during training.
Errors are computed over the training time-interval $T_{\text{train}} \subset T_{\text{test}}$ in the paragraphs discussing the quality of the approximation of individual trajectories, of clustering patterns and of the Basset force.
The two paragraphs discussing temporal generalization use errors computed over the longer test interval $T_{\text{test}}$.
\subsection{Error metrics}
We use the maximum point-wise relative distance from the UDE-approximated trajectories to the full MaRGE reference solution to assess accuracy.
Let $\mathbf{y_1}, \mathbf{y_2}, ..., \mathbf{y_n}$ with $\mathbf{y_i} \in \mathbb{R}^{3}$ be the particle positions in the reference trajectory and $\mathbf{\hat{y}_1}, \mathbf{\hat{y}_2}, ... , \mathbf{\hat{y}_n} $ with $\mathbf{\hat{y}_i} \in \mathbb{R}^{3}$ the positions in the approximate trajectory at the same time-steps.
The maximum relative point-wise distance then reads 
\begin{equation}
  d(\mathbf{\hat{y}}, \mathbf{y}) =\frac{ \max_{i \in {1,..., n}}\{|| \mathbf{\hat{y}_i} - \mathbf{y_i} ||_{2}\}}{\max_{i \in {1,..., n}}||\mathbf{y_i}||_{2}}.
  \label{eq:Evaluation Formula}
\end{equation}
We further consider minimum and average point-wise relative distances. 
For the clustering analysis, we calculate the relative distances
\begin{equation}
  d_\text{final}(\mathbf{\hat{y}}, \mathbf{y}) =\frac{ || \mathbf{\hat{y}_n} - \mathbf{y_n} ||_{2}}{||\mathbf{y_n}||_{2}}
  \label{eq:Evaluation Formula Final Positions}
\end{equation}
between the final positions of a UDE-trajectories and ground-truth references.
Finally, we compare the outputs of the networks against the Basset term $H(t)$ in~\cref{eq:MaRGE2} in the reference computation. 
Since $H(t)$ is the integrated Basset term, we integrate the output from the neural networks over all time-steps using the trapezoidal rule.
For this comparison, we omit the constant prefactor $-R\sqrt{\frac{3}{S \pi}}$. We denote the individual components of the integrated basset term by $H^{(1)}$, $H^{(2)}$, and $H^{(3)}$.

%
%
\subsection{Vortex Field}
In this subsection we analyze the accuracy of the UDE approximation for the vortex flow field.
We use the networks that have been trained with the full training data set containing 100 trajectories unless explicitly indicated otherwise.

\begin{figure}[th]
  \centering
  \includegraphics[width=\textwidth]{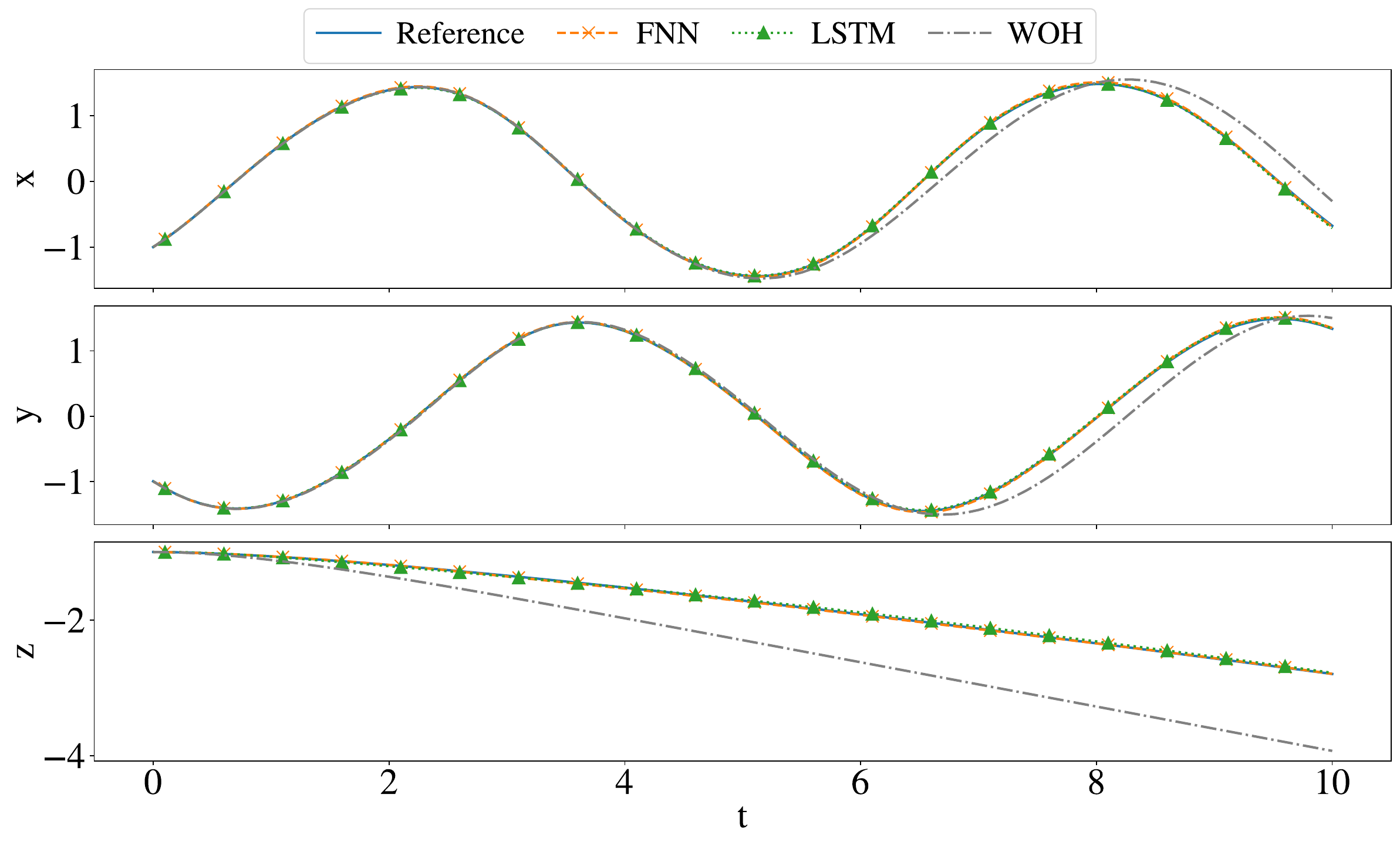}
  \caption{Trajectory components of the UDE approach with FNN and LSTM applied to an initial position unseen during training. The ground-truth, computed by solving the full MaRGE, is shown as solid line. For comparison, the trajectory ignoring the history term is shown as a dash-dotted line. In line with previous studies, ignoring the Basset force causes the particle to sink too fast under gravity~\cite{PrasathEtAl2019}. By contrast, the UDE-approximation provides the correct sinking rate.}
  \label{fig:Test Trajectories 2D Exp 01}
\end{figure}

\paragraph{Individual trajectories.}
\cref{tab:Trajectories exp01 tspan1} shows the maximum, minimum and average point-wise distance~\cref{eq:Evaluation Formula} between UDE-generated trajectories training on a data set with ten (middle column) and 100 trajectories (right column) and the ground-truth.
For comparison, distances to a trajectory that ignores the history term completely (WOH) are also shown. 
The three components of the positions of one example trajectory starting at an initial position from the test set are shown in \cref{fig:Test Trajectories 2D Exp 01}.
\begin{table}[th]
 \setlength{\tabcolsep}{10pt}
  \centering
  \begin{tabular}{@{}l  c  c  c  c  c@{}}
  	\toprule
               & WOH                 & FNN(5)              & LSTM(5)               & FNN(100)            & LSTM(100)           \\
    \midrule
    avg. dist. & $5.901\cdot10^{-1}$ & $2.316\cdot10^{-1}$ & $1.204\cdot10^{-1}$   & $7.034\cdot10^{-3}$ & $6.692\cdot10^{-3}$ \\
    min. dist. & $3.797\cdot10^{-1}$ & $ 1.418\cdot10^{-2}$ & $1.144\cdot10^{-2}$   & $1.545\cdot10^{-3}$ & $1.458\cdot10^{-3}$ \\
    max. dist. & $1.133\cdot10^{0}$ & $8.363\cdot10^{-1}$  & $4.761\cdot10^{-1}$ & $2.418\cdot10^{-2}$ & $2.033\cdot10^{-2}$ \\ \bottomrule
  \end{tabular}
  \caption{Errors of the UDE-computed trajectories in the test set for a training set of five (middle) and 100 (right) trajectories. The WOH column shows the errors that are achieved by simply neglecting the Basset force. For a small training dataset, both models are slightly more accurate than ignoring the history term completely, with the LSTM showing a small advantage over the FNN. For the larger training set, both FNN and LSTM produce similar accuracy and are about two orders of magnitude more precise than the reference without the history term.}
  \label{tab:Trajectories exp01 tspan1}
\end{table}

The results show that the LSTM performs better when there are fewer trajectories in the training set but that it loses this advantage as the amount of training data increases. 
With a training set of 100 trajectories, both networks perform equally well on the unseen data and are two orders of magnitude more accurate than simply neglecting the history term.

\begin{figure}[t]
  \centering
  \includegraphics[width=\textwidth]{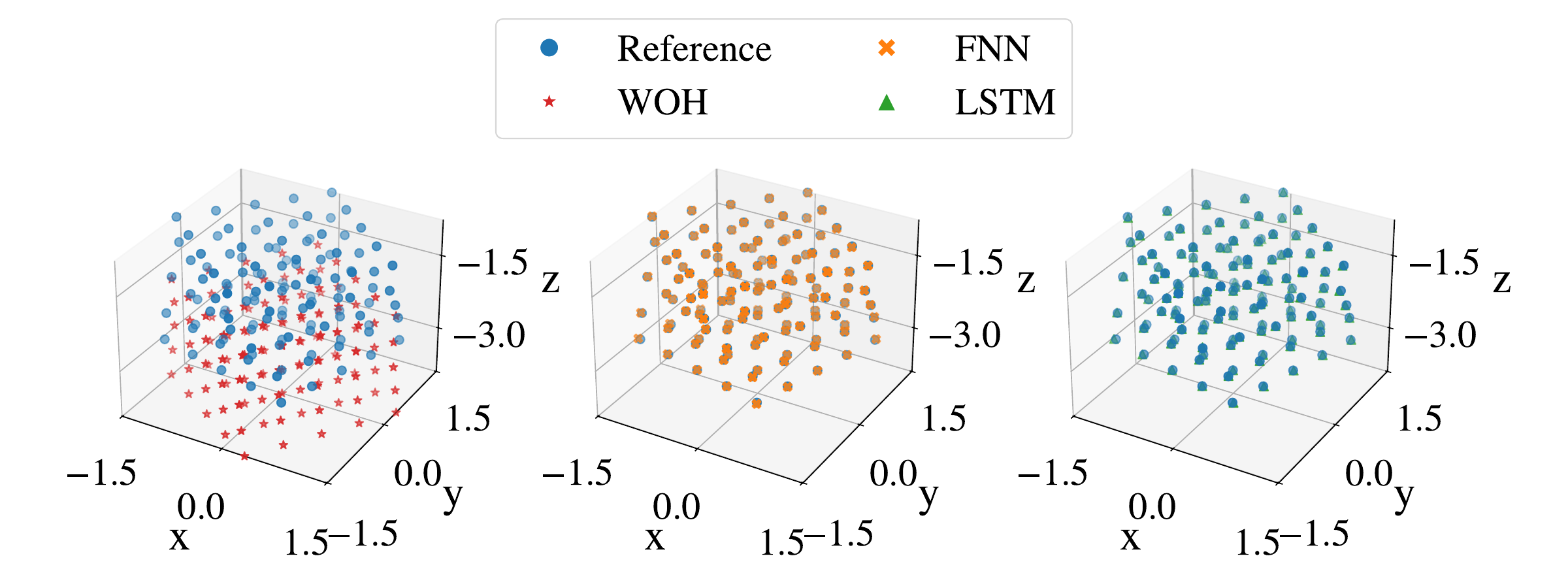}
  \caption{Clustering of particles at $t=10$ in the vortex. Ignoring the Basset force leads to an overestimation of the sinking speed of particles and thus a qualitatively different clustering pattern. Both FNN and LSTM are able to produce the correct patterns.}
  \label{fig:Clustering_vortex}
\end{figure}

%
\paragraph{Clustering patterns.}
In many applications of MaRGE, the main concern is not necessarily individual trajectories but rather the larger patterns that are generated by the motion of many particles~\cite{Urizarna2025b}.
We compare how particles cluster at time $t = 10$ when their movement is modelled using full MaRGE, the UDE approximation of the Basset force and when the history term is ignored.
The resulting patterns for the vortex field are shown in~\cref{fig:Clustering_vortex} while~\cref{tab:Clustering Results Vortex} shows the minimum, maximum and average distance of the final particle positions.
Both networks provide optically close approximations of the true clustering pattern.
The average distances are $5.667\cdot10^{-3}$ for the LSTM and $6.256\cdot10^{-3}$ for the FNN, which is 2 orders of magnitude lower than the solution that neglects the history term with an average distance of $6.066\cdot10^{-1}$.
Overall, both network architectures are capable of approximating the Basset term well enough such that the clustering patterns closely match those from the reference solution.

\begin{table}[th]
\setlength{\tabcolsep}{10pt}
\centering
\begin{tabular}{l  c  c  c}
	\toprule
             & WOH                 & FNN(100)            & LSTM(100)           \\
  \midrule
  avg. dist. & $6.066\cdot10^{-1}$ & $6.256\cdot10^{-3}$ & $5.667\cdot10^{-3}$ \\
  min. dist. & $3.797\cdot10^{-1}$ & $6.784\cdot10^{-4}$ & $4.791\cdot10^{-4}$ \\
  max. dist. & $1.425\cdot10^{0}$  & $2.737\cdot10^{-2}$  & $1.961\cdot10^{-2}$ \\
  \bottomrule
\end{tabular}
  \caption{Average, minimum, and maximum distances of the final positions in the clustering plot~\cref{fig:Clustering_vortex}. Ignoring the Basset force leads to substantial inaccuracies in the formed patterns because particles sink too fast. Both FNN and LSTM improve the accuracy of the particles' final positions by two orders of magnitude.}
  \label{tab:Clustering Results Vortex}
\end{table}

\begin{figure}[t]
  \centering
  \includegraphics[width=\textwidth]{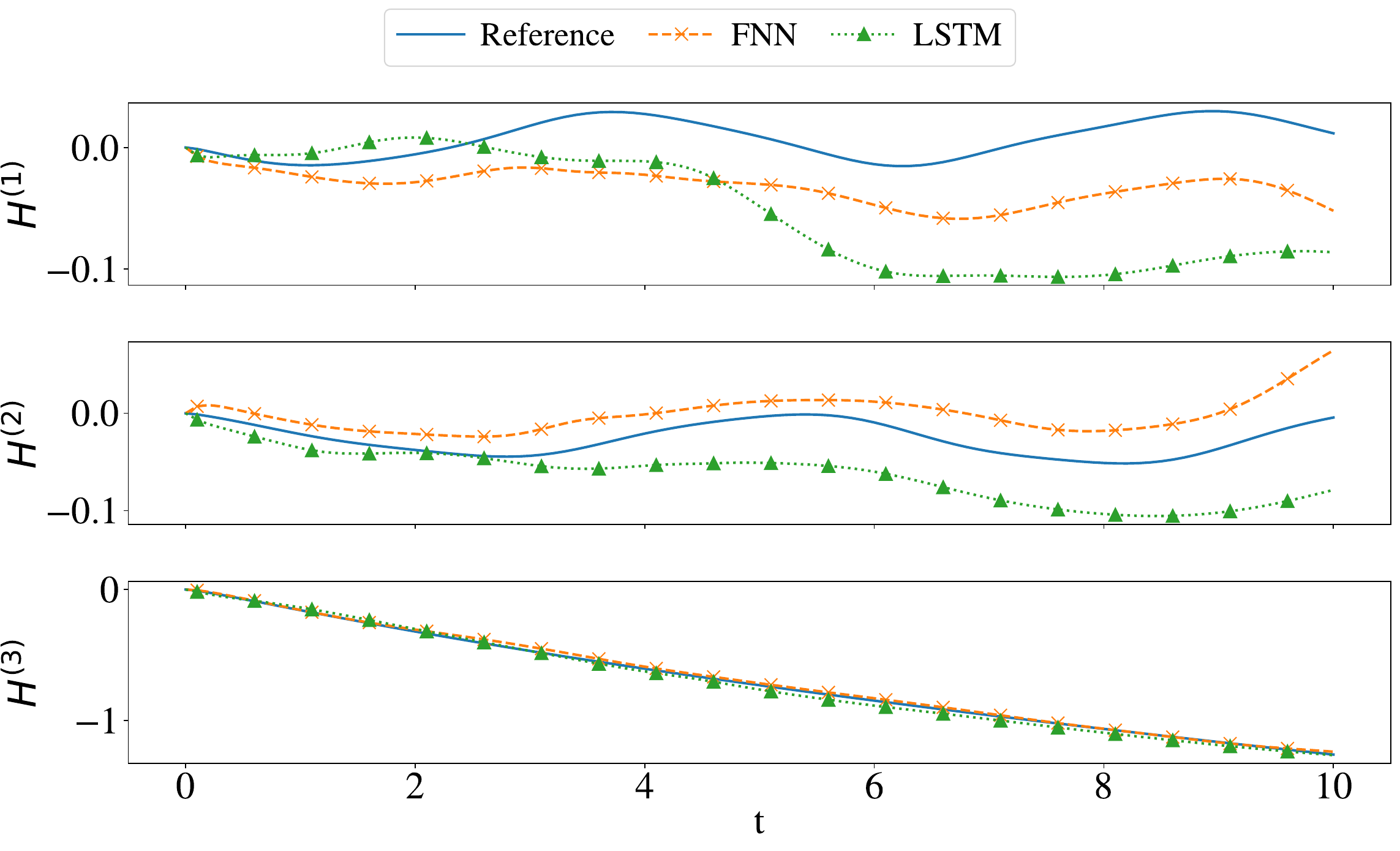}
  \caption{The three components of the Basset force in the reference simulation and approximations provided by the UDE approach over time. Note the differences in scale on the vertical axes. Since the impact of the Basset force is most important in the vertical, the training seems to prioritize accuracy in the vertical component, leading to the correct sinking speed.}
  \label{fig:Basset Comparison Exp 01}
\end{figure}

\paragraph{Basset force.}
The three components of the integrated Basset term~\cref{eq:H_calc} and their approximation by FNN and LSTM over time are shown in~\cref{fig:Basset Comparison Exp 01}.
Both networks produce a close approximation in the z-coordinate, but show larger deviations from the reference in the horizontal coordinates.
They capture the oscillating behaviour, but are unable to match frequency or amplitude.
The reason is the difference in scale between the horizontal and vertical Basset force components.
In the z-coordinate, the force is of order unity while its horizontal components are of the order of $0.01$.
Therefore, from the view of the network, the errors in the horizontal components have a small quantitative effect on the resulting trajectory and training focusses on producing the correct vertical Basset force.

\begin{figure}[ht]
  \centering
  \includegraphics[width=\textwidth]{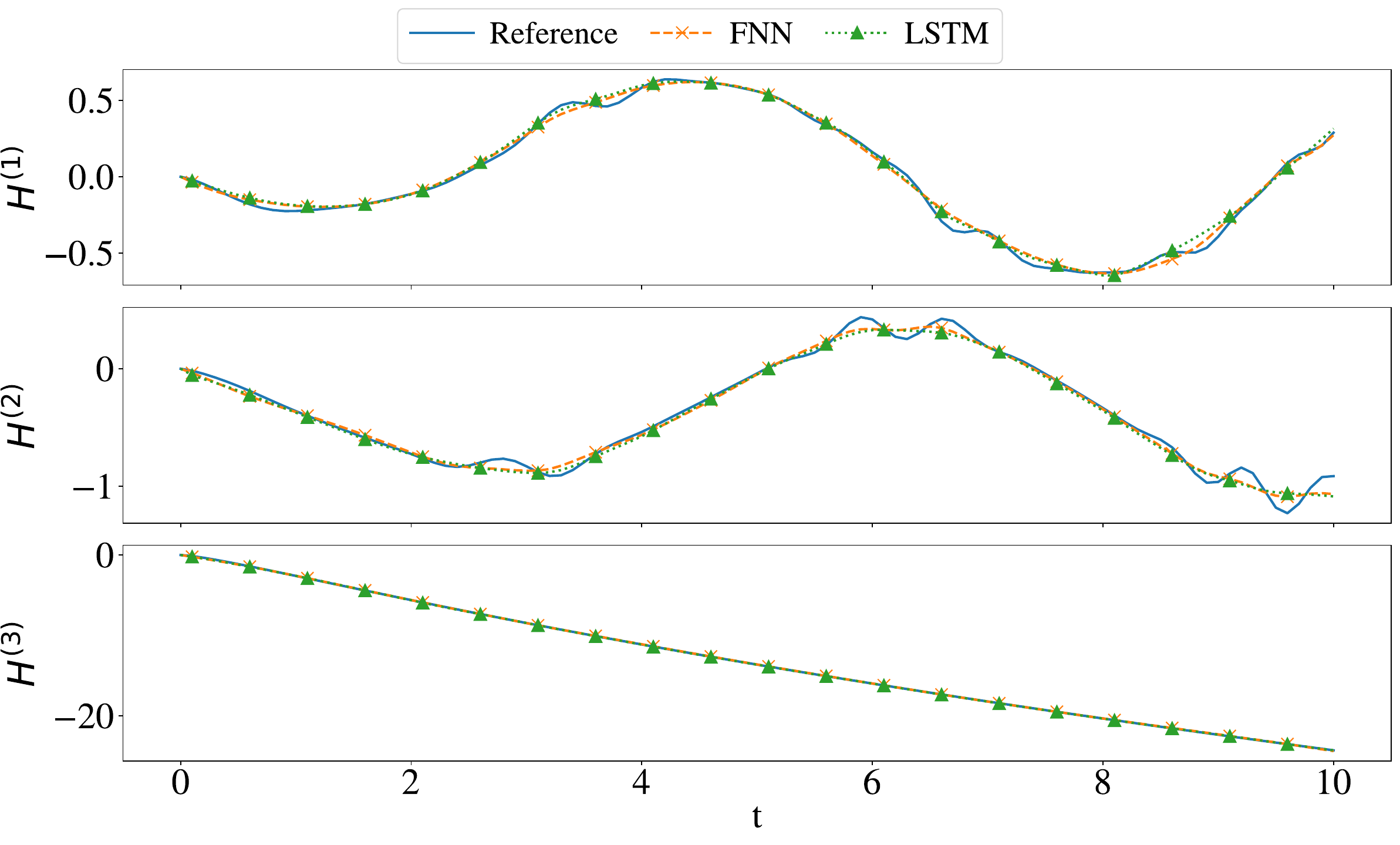}
  \caption{Approximation of the Basset force with increased particle density. The networks now correctly approximate the horizontal components of the Basset force, as these are now having the most important impact on particle motion.}
  \label{fig:Basset Overfitting}
\end{figure}

To confirm that the UDE can also correctly capture the horizontal Basset forces, we perform an additional experiment.
For simplicity, we use only a single trajectory for training, and increase particle density so that $R=0.6$ to increase the effect of the history term on horizontal motion.
The resulting Basset forces are shown in~\cref{fig:Basset Overfitting}. 
We now observe a very good match of the horizontal components of the Basset force, demonstrating that the network correctly identifies the physical dimensions where it has the strongest influence.

\begin{figure}[th]
  \centering
  \includegraphics[width=\textwidth]{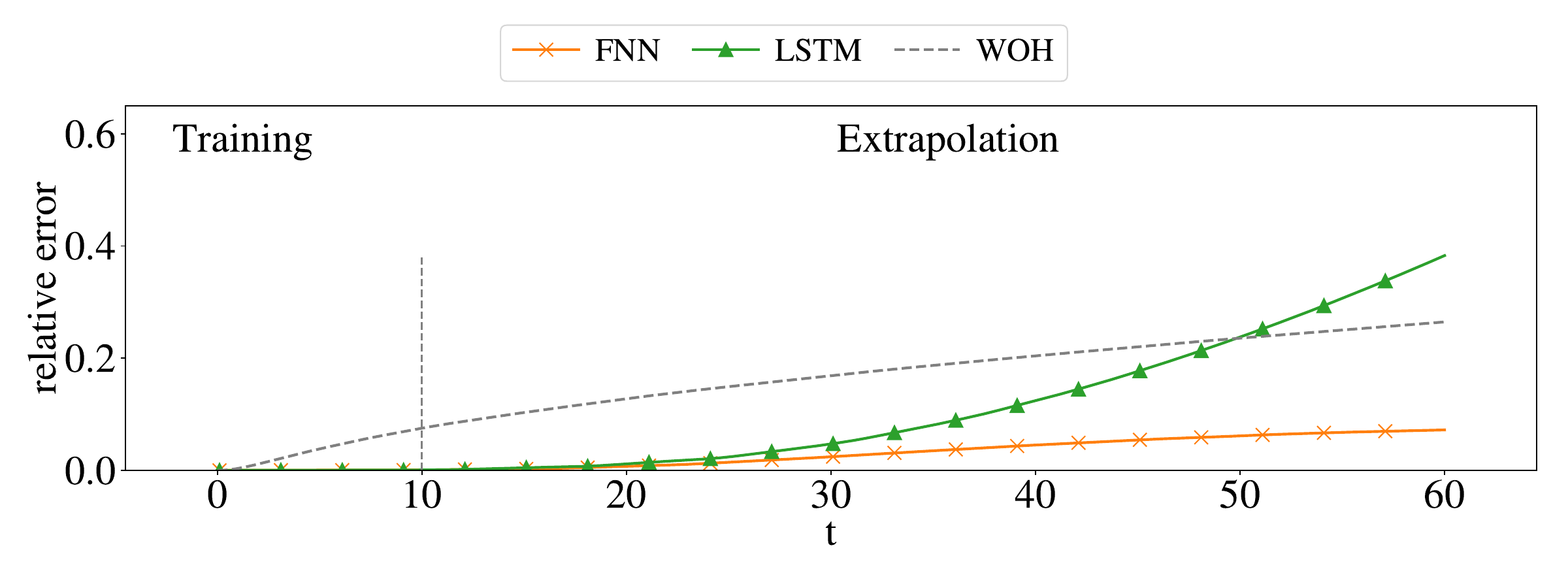}
  \caption{Relative errors of the UDE models with the solution that ignores the history term shown as reference for the vortex field. 
  The networks generalize beyond the training interval but not indefinitely. Starting at $t=20$  the error starts to grow. However, even at $t=60$ the FNN remains substantially more accurate than ignoring the Basset force.}
  \label{fig:Error over time vortex}
\end{figure}

\paragraph{Generalization in time.}
To see how well the UDE generalizes beyond the training time interval, we plot the relative positional error against time over $T_{\text{test}}$ in~\cref{fig:Error over time vortex}.
During the training timespan $T_{\text{Train}}$, both networks produce small relative errors in line with the analysis before.
By contrast, ignoring the history term produces a roughly linear increase in error.
However, both models eventually reach the limits of their generalization capability.

Both FNN and LSTM maintain a small,  very slowly growing error until $t=20$. 
After that, the error also starts to grow more quickly for both, although much slower for the FNN than for the LSTM.
At time $t=60$, six times beyond the training interval, FNN is still substantially more accurate than ignoring the Basset force.
In summary, while the networks cannot model the Basset force indefinitely, they generalize substantially beyond their training timespan.
Further tests not shown here suggest that the error growth is sensitive to the accuracy of the underlying ODE solver. 
More accurate solvers did generalize better, but at the cost of longer training times. 
Also, the UDE's generalization capabilities seem to change between multiple runs using the same experimental settings.  
A more detailed investigation of the impact of discretization errors on the performance of UDEs is left for future work.

\begin{figure}[ht]
  \centering
  \includegraphics[width=\textwidth]{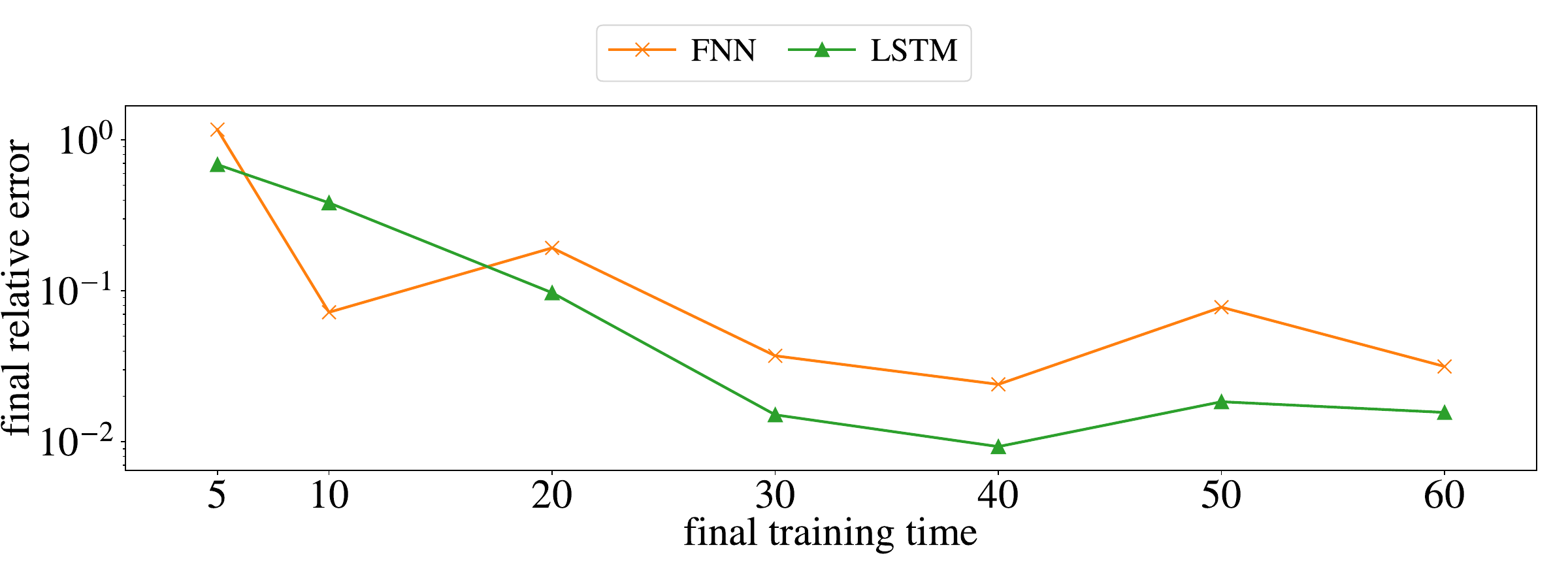}
  \caption{Average final relative error on the test set versus the length of the training interval $T_{\text{train}}$. The testing interval is always 50 time units longer than the training interval, with final times ranging from $55$ to $110$.}
  \label{fig:Final Relative Errors}
\end{figure}

To investigate how the generalization capability of the network depends on the length of the training interval, train on intervals of varying length and then extrapolate over an interval that is $50$ nondimensional time units longer.
We start with $T^{(1)}_{\text{train}} = [0.0, 5.0]$ and $T^{(1)}_{\text{test}} = [0.0, 55.0]$ and end at $T^{(7)}_{\text{train} = [0.0, 60.0]}$,  $T^{(7)}_{\text{test}} = [0.0, 110.0]$. 
We use 100 trajectories in the training set and use the same parameters as in the vortex experiment ~\cref{tab:Parameter MaRGE Vortex field}. In particular, 
the second experiment with $T^{(2)}_{\text{train}} = [0.0, 10.0]$ and $T^{(2)}_{\text{test}} = [0.0, 60.0]$ corresponds exactly to the experiment with the training set of 100 trajectories. 
\cref{fig:Final Relative Errors} shows the average final relative error on the testing set against training time.
Final errors decrease until a training interval of $[0.0, 40.0]$ and then increase again. 
This might suggest that the lowest errors are achieved if training and extrapolation time are roughly the same, but a detailed investigation is left for future work. 

\subsection{Experimental field}
This subsection repeats the investigation of individual trajectories, clustering patterns, approximations of the Basset force and generalization in time for the experimental flow field.

\paragraph{Individual trajectories}
The analysis is carried out analogously to the vortex field. 
The resulting errors are shown in~\cref{tab: Trajectories exp02 tspan1}.
Both networks produce about one order of magnitude smaller errors than ignoring the Basset force.
Gains are somewhat less than for the vortex, likely because the particles do not sink as far due to the shorter time interval and because of sharper gradients in the flow field.

\begin{figure}[t]
   \centering
   \includegraphics[width=\textwidth]{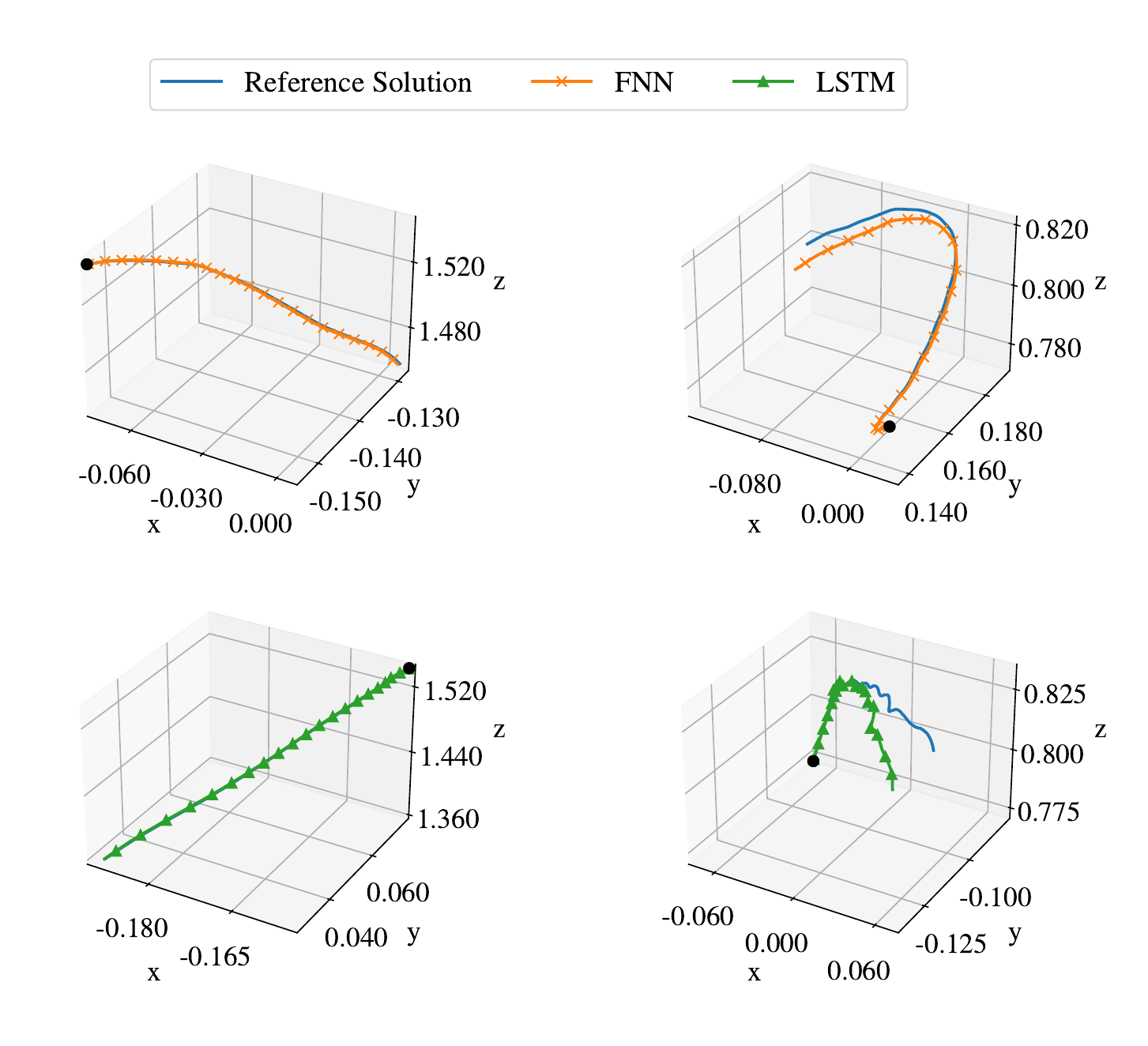}
   \caption{Trajectories produced by the UDE approximation and reference solution for the experimental flow field. The top figures show the best (left) and worst (right) approximation by the FNN while the lower figures show the best (left) and worst (right) approximation from the LSTM. Both worst-cases show a very sharp turn, suggesting that rapid changes of direction of the particle might present a challenge to the UDE approach.}
   \label{fig:Test Trajectories 3D Exp 02}
\end{figure}

\begin{figure}[t]
  \centering
  \includegraphics[width=\textwidth]{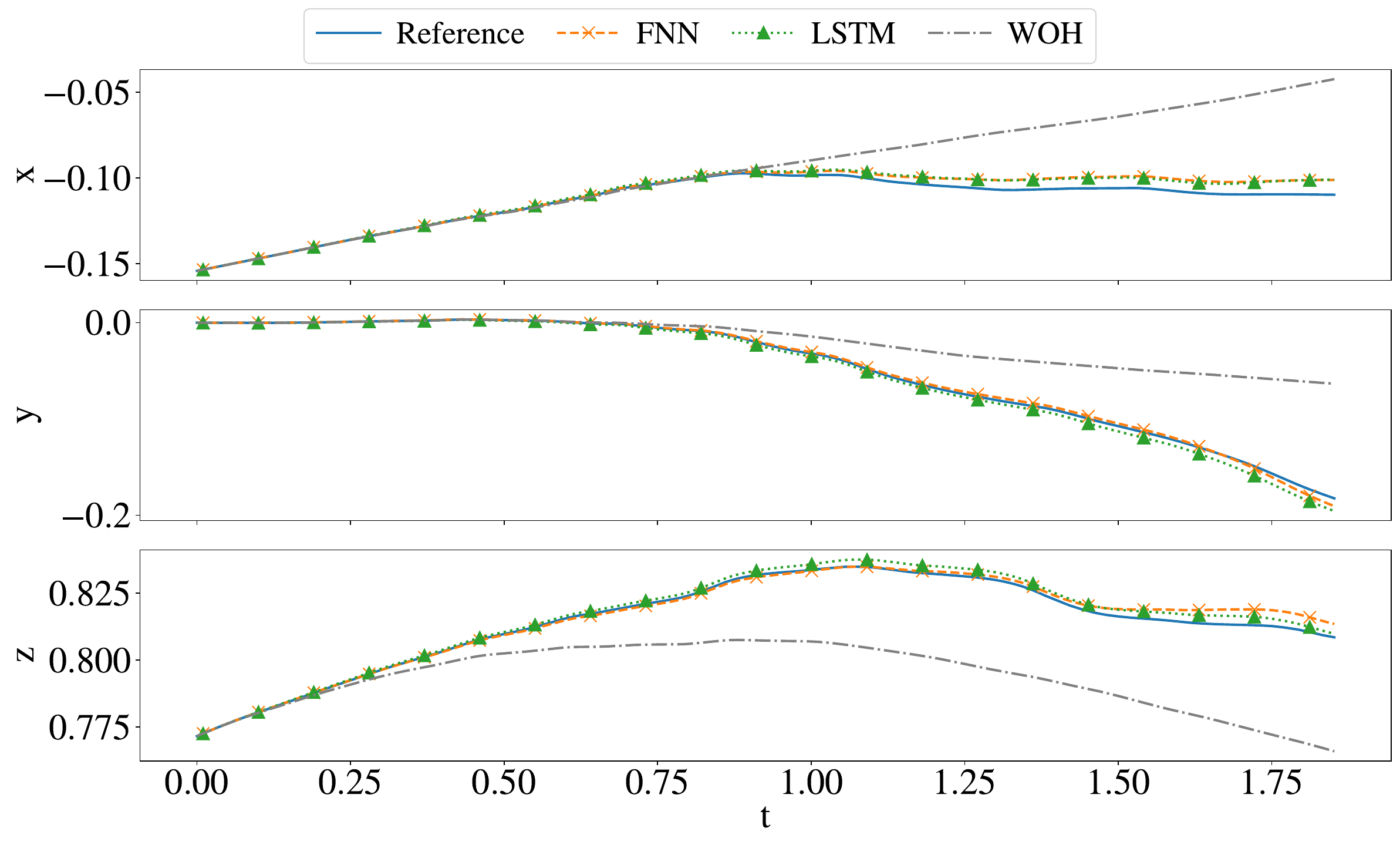}
  \caption{Example trajectory for an initial position from the test split into coordinates. Both networks are able to approximate the solution with reasonable accuracy while ignoring the history term leads to substantial errors in all three directions.}
  \label{fig:Test Trajectories 2D Exp02}
\end{figure}

The trajectories responsible for the maximum and minimum error for FNN and LSTM are shown in~\cref{fig:Test Trajectories 3D Exp 02}. 
Note how the best-case trajectories on the left are mostly straight lines while the worst-cases feature sharp turns.
Inertial effects and thus the relevance of the Basset force are greater for such turns, so that approximation errors from the networks translate into larger positional differences.
\cref{fig:Test Trajectories 2D Exp02} shows the three components of the positions of one example trajectory starting at an initial position from the test set.
While the errors in the UDE generated positions are larger than for the vortex, they remain substantially smaller in all three directions than when the history term is ignored.

%
%

\begin{table}[t]
\setlength{\tabcolsep}{10pt}
  \centering
    \begin{tabular}{l c  c c } \toprule
                & WOH               &  FNN                & LSTM              \\
      \midrule
      avg. dist.&$8.312\cdot10^{-2}$& $3.468\cdot10^{-3}$ & $4.876\cdot10^{-3}$ \\
      min. dist.&$5.048\cdot10^{-2}$& $3.934\cdot10^{-4}$  & $2.066\cdot10^{-4}$ \\
      max. dist.&$1.1\cdot10^{-1}$& $1.744\cdot10^{-2}$ & $2.758\cdot10^{-2}$ \\ \bottomrule
    \end{tabular}
  \caption{
    Average, minimum, and maximum distances to the reference solutions for both models over the timespan $T_{\text{train}} = [0, 1.852]$ for the experimental field. The UDE approach reduces errors by about one order of magnitude compared to ignoring the Basset force.}
  \label{tab: Trajectories exp02 tspan1}
\end{table}

%
%

%
%

\begin{figure}[t]
  \centering
  \includegraphics[width=\textwidth]{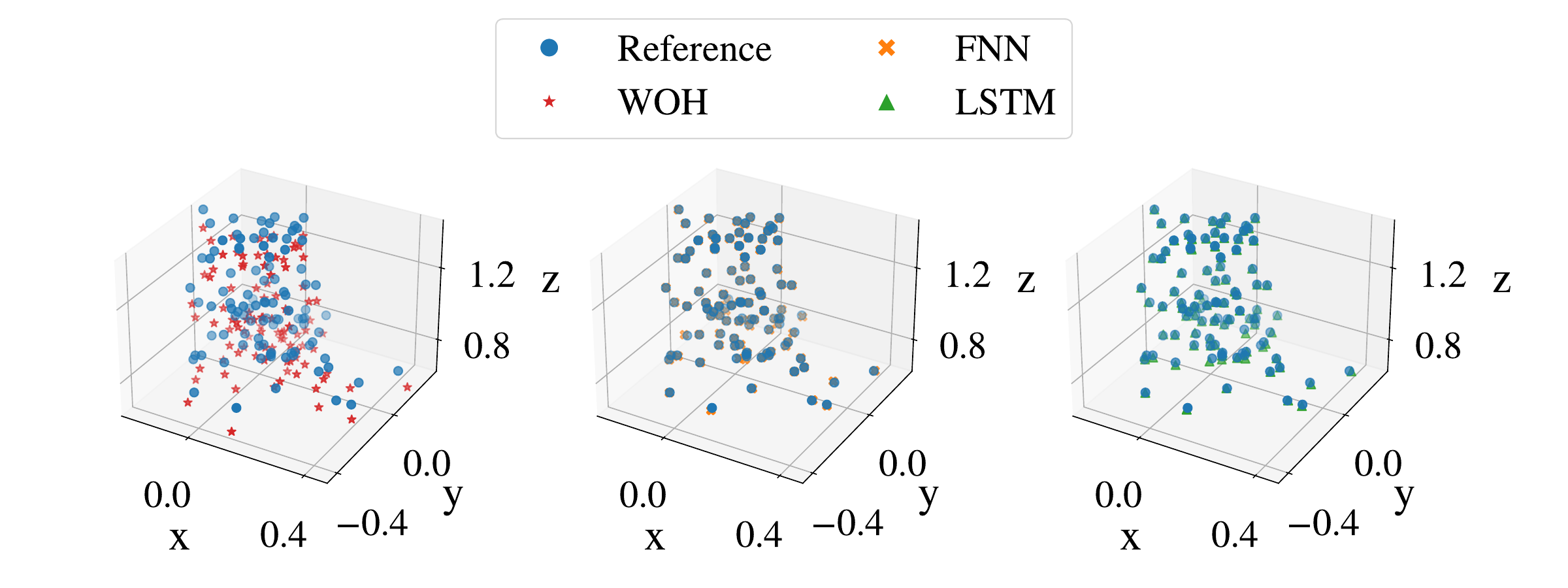}
  \caption[Final position clustering experimental field.]{Clustering of the final positions for both networks. The blue dots represent the reference solution, the red stars the solution that neglects the history term (WOH), the orange crosses the UDE with the FNN model, and the green triangles the UDE with the LSTM model. 
  Like on the vortex field, the WOH solution overestimates the particle movement in the z-direction, which leads to an offset of the final positions. The UDE is able to compensate for this.}
  \label{fig:Clustering Experimental Field}
\end{figure}

\paragraph{Clustering.}
Clustering patterns are shown in~\cref{fig:Clustering Experimental Field} while~\cref{tab:clustering results exp02} shows the distances between final particle positions.
As for the vortex, ignoring the Basset force produces a visibly different pattern while both FNN and LSTM produce patterns that are optically identical.
Both models reduce the average distance by one order of magnitude, again slightly less than what was observed for the vortex.
The FNN slightly outperforms the LSTM in average and maximum distance, although the difference is small.
\begin{table}[t]
 \setlength{\tabcolsep}{10pt}
  \centering
  \begin{tabular}{l c  c c } \toprule
                & WOH   & FNN                 & LSTM                \\
      \midrule
      avg. dist.& $9.653\cdot10^{-2}$ & $3.814\cdot10^{-3}$& $5.307\cdot10^{-3}$   \\
      min. dist.&$5.993\cdot10^{-2}$ & $4.178\cdot10^{-4}$ & $1.913\cdot10^{-4}$   \\
      max. dist. &$2.368\cdot10^{-1}$ & $1.828\cdot10^{-2}$ & $2.773\cdot10^{-2}$  \\
      \bottomrule
    \end{tabular}
  \caption{
    The average, minimum, and maximum distances of the final positions of the UDEs and the solution that neglects the history term (WOH) to the final positions of the reference solution for both models.
    Both models lower the average distance by one order of magnitude.}
  \label{tab:clustering results exp02}
\end{table}
\newpage

\begin{figure}[ht]
  \centering
  \includegraphics[width=\textwidth]{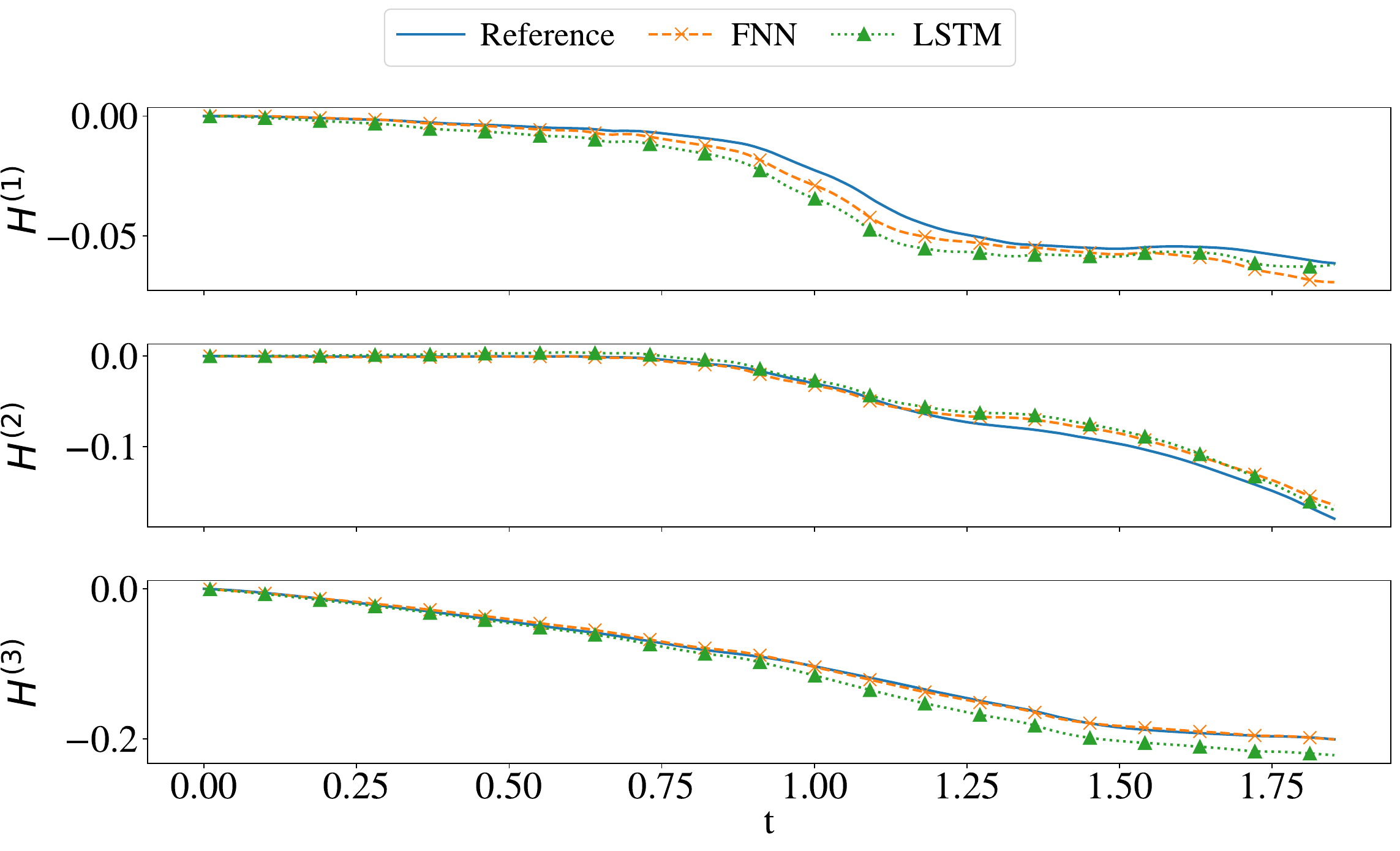}
  \caption{Components of the Basset force integrated over time for the experimental field. Both models approximate the Basset force correctly in all coordinates.}
  \label{fig:Basset Comparison Exp 02}
\end{figure}

\paragraph{Basset force.}
The three components of the Basset force for the experimental field, integrated over time, are shown in~\cref{fig:Basset Comparison Exp 02}. 
The UDE produces a good approximation of the Basset term in all coordinates. This indicates that the structure of the Basset term has been learned correctly from the data.

\begin{figure}[t]
  \centering
  \includegraphics[width=\textwidth]{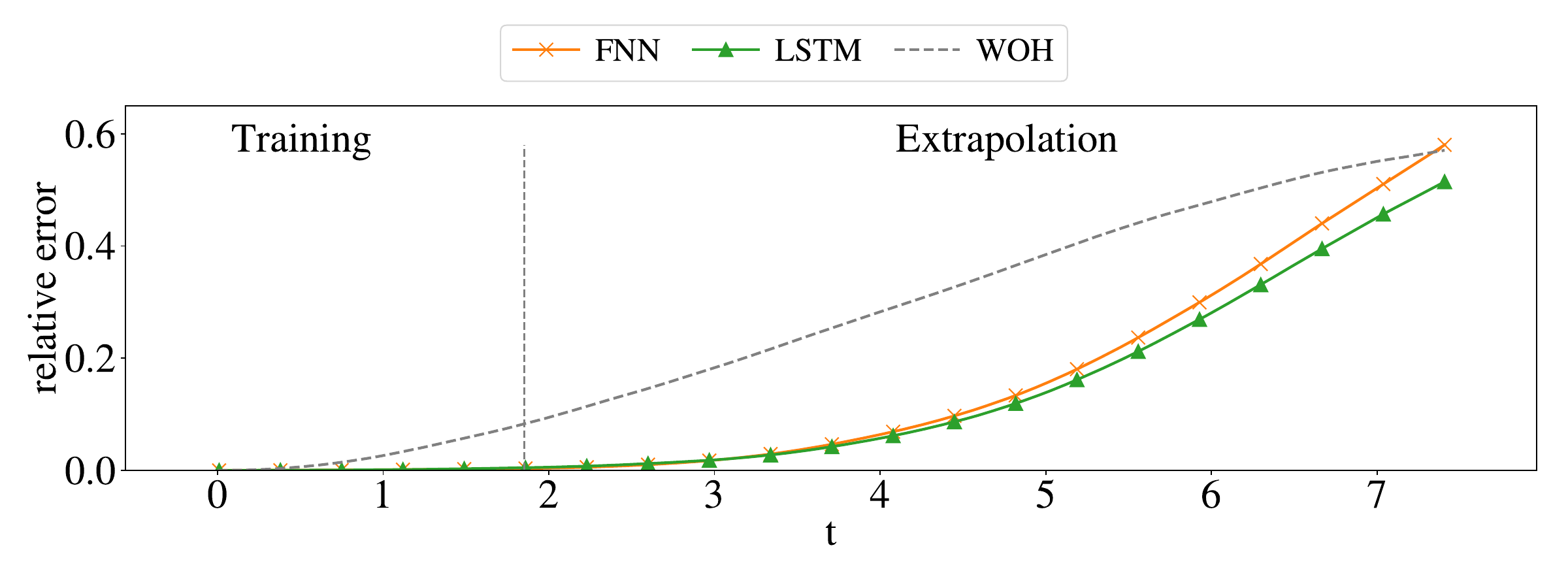}
  \caption{Error over time for the experimental field. 
  Both models keep the relative error below the solution that neglects the history term for a time longer than the training time, but the error increases with time. 
  At the end of the testing timespan $T_{\text{Test}}$, the FNN surpasses the WOH solution.}
  \label{fig:Error over time Experimental}
\end{figure}

\paragraph{Generalization in time.}
Errors over $T_{\text{test}}$ are shown in~\cref{fig:Error over time Experimental}. 
The results are comparable to the vortex field, but both FNN and LSTM stop generalizing at roughly the same time $t=4$.
Errors for the LSTM increase more slowly than for FNN and remain below the reference ignoring the Basset force until the end.
Shortly before $t=7$, the FNN starts producing larger errors than ignoring the history term, so care must be taken in applications to not exceed the time interval used for training by too much.
Our results suggest that generalizing in time up to three times the training interval should produce reliable results, although strongly turbulent flow fields might require even stricter safety margins.

%
%

\begin{figure}[ht]
  \centering
  \includegraphics[width=\textwidth]{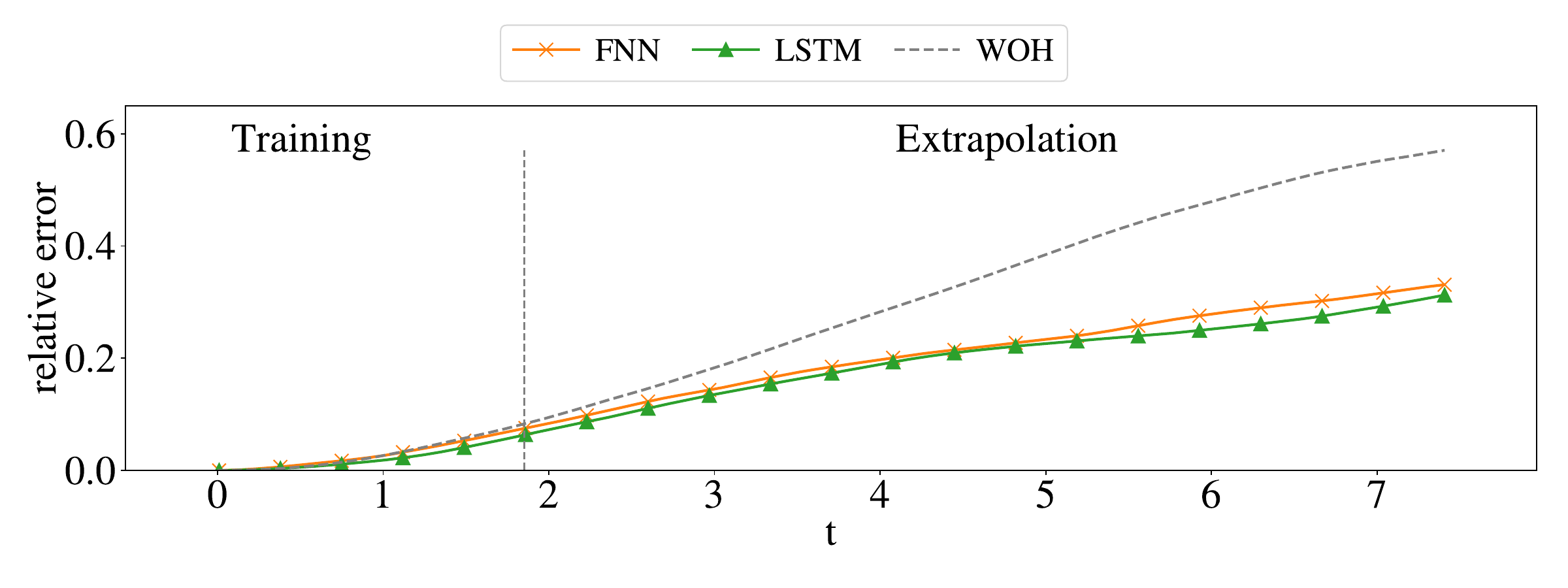}
  \caption{Error over time for the experimental field with networks that were trained on the vortex field.}
  \label{fig:Error over time exchanged network weights}
\end{figure}

\paragraph{Generalization to different flow fields.}
So far, in all cases we considered, the UDE was applied to the same fluid field that it was trained on.
However, the Basset term in~\cref{eq:MaRGE2} only depends on the relative velocity $\mathbf{w}$ and not explicitly on the flow-field $\mathbf{u}$.
Therefore, the structure that the network learns from one flow-field should remain valid for another.
To assess how well the UDE generalizes to unseen flow-fields, we apply the network trained on the vortex field to the experimental velocity field. 
\cref{fig:Error over time exchanged network weights} shows the resulting error over time. 
Compared to the error for the network trained on the experimental field see~\cref{fig:Error over time Experimental}, for both models, 
the error is surprisingly even smaller than when the network is trained on the experimental flow field.
Presumably this is because the much simpler structure of the vortex makes it easier for the network to learn the structure of the Basset force term.
However, a more detailed investigation of this effect is left for future work.
In summary, these results suggest that the UDE approach generalizes very well across flow-fields and that the network can be redeployed to unseen flows without much or any retraining.
This is important in applications where high-fidelity data about the flow-field might not always be available.

%
%
%
%
%
%
%
%
\section{Conclusion}
We show a new approach for approximating the Basset force in the Maxey-Riley-Gatignol equations (MaRGE), using the concept of universal differential equations.
This provides a data-based approximation and the resulting system of ordinary differential equations can be easily solved using standard libraries.
The approach is tested on two different fluid flow fields, an analytically given three-dimensional vortex and flow-field in a lab-scaled stirred tank reactor interpolated from experimental data.
In both cases, the UDE approximation produces a good approximation of the Basset history term and leads to substantially higher accuracies than simply neglecting it.
We also show that the UDE approximation leads to realistic sinking and clustering patterns, important qualitative features that are incorrect when the Basset force is not taken into account.
The approach also generalizes well in time and networks can provide good accuracy even for flow-fields not seen in training.
Our results suggest that data-based approximations of the integral term in MaRGE are promising way to include its important effects in situations where a full numerical solution might not be desired or feasible.

\section*{Acknowledgments}
We would like to thank Eike Steuwe and Prof.~Dr.~Alexandra von Kameke at the Hamburg University of Applied Sciences (HAW Hamburg) for providing the data for the stirred tank reactor flow field.

~\clearpage
\section*{References}
\bibliography{refs}

\end{document}